\documentclass{article}

\PassOptionsToPackage{numbers, compress}{natbib}
\usepackage[preprint]{neurips_2026}
\usepackage{tabularx}
\usepackage{algorithm}
\usepackage{algorithmic}
\usepackage{subcaption}
\usepackage{svg}
\usepackage{microtype}
\usepackage{graphicx}
\usepackage{booktabs} 
\usepackage{longtable}
\usepackage{siunitx}
\usepackage{caption}
\usepackage{xcolor}
\usepackage{bm}
\usepackage{multirow}
\usepackage{float}

\usepackage[utf8]{inputenc} 
\usepackage[T1]{fontenc}    
\usepackage{hyperref}       
\usepackage{url}            
\usepackage{booktabs}       
\usepackage{amsfonts}       
\usepackage{nicefrac}       
\usepackage{microtype}      
\usepackage{xcolor}         

\usepackage{amsmath}
\usepackage{amssymb}
\usepackage{mathtools}
\usepackage{amsthm}
\usepackage{wrapfig}
\usepackage[capitalize,noabbrev]{cleveref}
\theoremstyle{plain}

\theoremstyle{definition}

\theoremstyle{remark}

\title{FastDSAC: Unlocking the Potential of Maximum Entropy RL in High-Dimensional Humanoid Control}

\author{%
  Jun Xue\textsuperscript{1, 2} \quad
  Junze Wang\textsuperscript{1, 3} \quad
  Shanze Wang\textsuperscript{1} \quad
  Xinming Zhang\textsuperscript{1} \quad
  Yanjun Chen\textsuperscript{1} \quad
  Wei Zhang\textsuperscript{1}\thanks{Corresponding author.} \\
  \\
  \textsuperscript{1}College of Information Science and Technology, Eastern Institute of Technology, Ningbo, China \\
  \textsuperscript{2}School of Computer Science, Shanghai Jiao Tong University, Shanghai, China \\
  \textsuperscript{3}College of Computer Science and Technology, China University of Petroleum(East China), Qingdao, China \\
  \texttt{zhw@eitech.edu.cn}
}

\begin{document}

\maketitle

\begin{abstract}
Scaling Maximum Entropy Reinforcement Learning (RL) to high-dimensional humanoid control remains a fundamental challenge, as the ``curse of dimensionality'' induces severe exploration inefficiency and training instability. Consequently, highly optimized deterministic policy gradients currently dominate high-throughput regimes. We address this limitation with FastDSAC, a framework that effectively unlocks the potential of maximum entropy stochastic policies for complex continuous control. We introduce Dimension-wise Entropy Modulation (DEM) to dynamically redistribute the exploration budget, alongside a continuous distributional critic tailored to ensure accurate value estimation by mitigating both high-dimensional overestimation and discrete quantization artifacts. Extensive evaluations on HumanoidBench and a diverse set of continuous control tasks demonstrate that FastDSAC establishes state-of-the-art performance for high-dimensional stochastic policies on the evaluated benchmarks. Our method is competitive with and often outperforms strong deterministic baselines, with gains of 180\% and 350\% on the challenging \textit{Basketball} and \textit{Balance Hard} tasks, respectively.
\end{abstract}


\begin{figure}[h]
    \centering
    \begin{subfigure}[b]{0.37\textwidth}
        \centering
        \includegraphics[width=\linewidth]{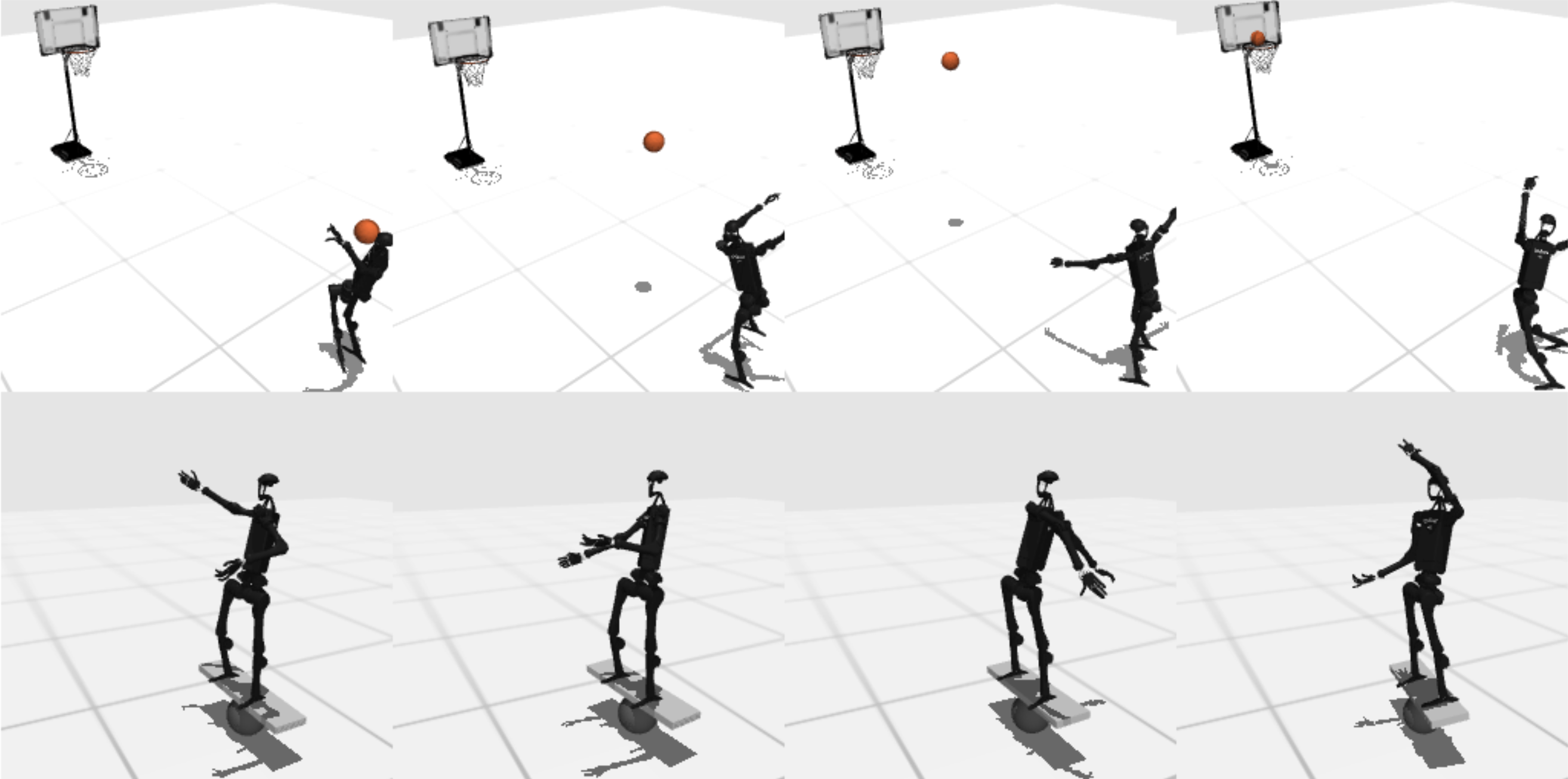}
        \caption{FastDSAC's performance on tasks}
        \label{fig:tasks_viz}
    \end{subfigure}
    \hfill
    \begin{subfigure}[b]{0.3\textwidth}
        \centering
        \includegraphics[width=\linewidth]{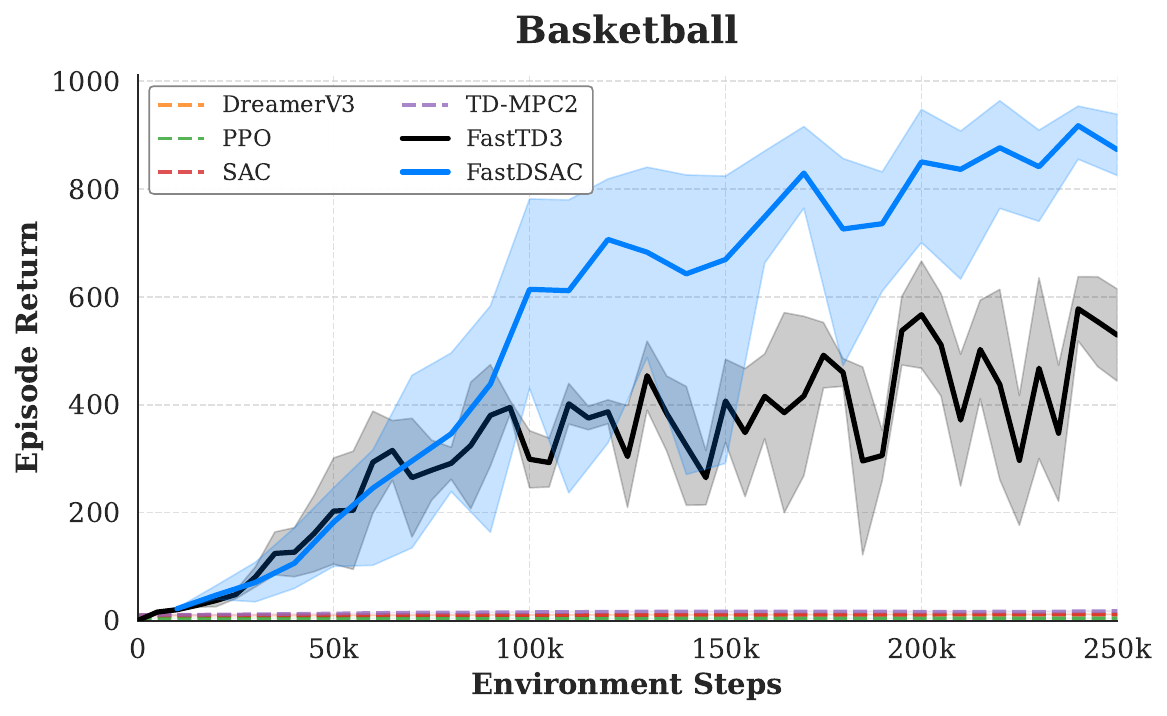}
        \caption{Basketball}
        \label{fig:basketball}
    \end{subfigure}
    \hfill 
    \begin{subfigure}[b]{0.3\textwidth}
        \centering
        \includegraphics[width=\linewidth]{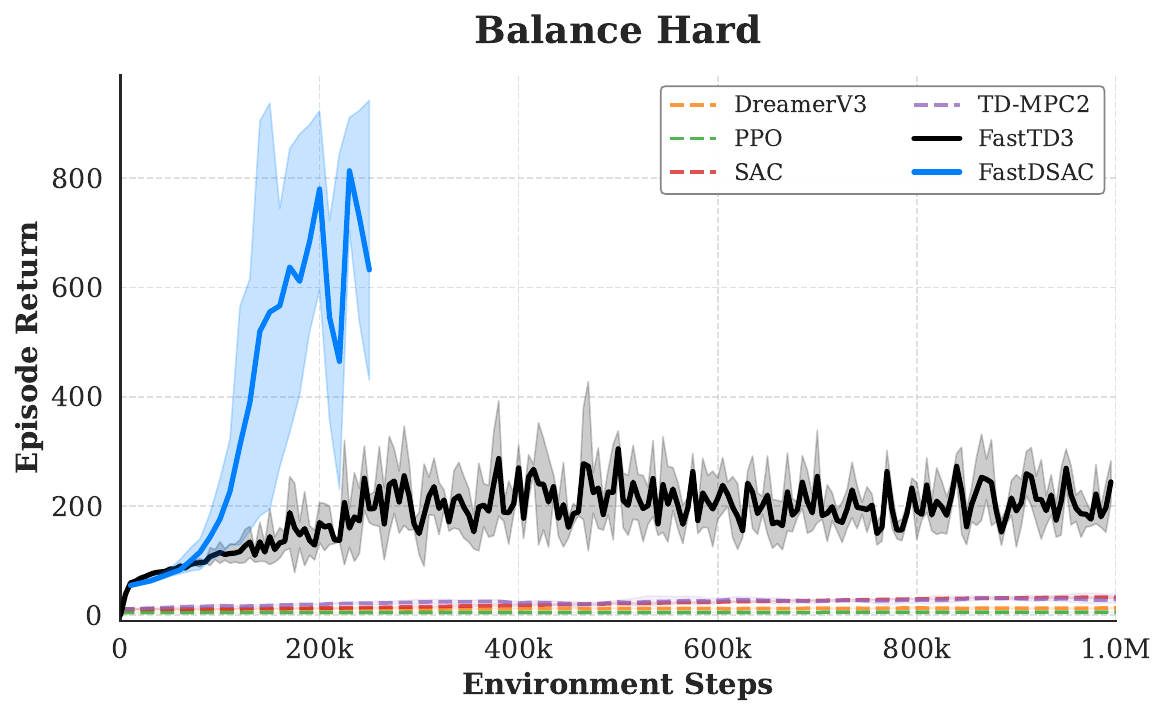}
        \caption{Balance Hard}
        \label{fig:balance_hard}
    \end{subfigure}
    \caption{\textbf{Performance of FastDSAC on very high-dimensional humanoid control tasks.} (a) Visualizations of the challenging \textit{Basketball} and \textit{Balance Hard} environments. (b, c) Evaluation curves comparing
  FastDSAC with the FastTD3 baseline. Curves show the mean over 5 seeds, with shaded regions spanning the min-max range across seeds. FastDSAC outperforms FastTD3, achieving final returns 1.8$\times$ and 3.5$ \times$ higher on the respective tasks with stronger sample efficiency.}
    \label{fig:combined_overview}
\end{figure}
\section{Introduction}
\label{sec:intro}
Deep reinforcement learning (RL) has driven significant progress in high-dimensional robotics and has become a practical framework for learning sensorimotor skills in humanoid control \citep{hansen2023td,hafner2023mastering,sferrazza2024humanoidbench,zhuang2025tdmpbc,seo2025fasttd3}. 
As morphological complexity and task demands increase, the field has shifted towards leveraging massively parallel simulation to accelerate training \citep{rudin2022learning,kaufmann2023champion,li2023parallel,shukla2025fastsac,gallicisimplifying}. 
In this high-throughput setting, deterministic policy-gradient methods, especially FastTD3 \citep{seo2025fasttd3}, have largely defined the state of the art. By leveraging the stability of deterministic policies in massively parallelized environments (e.g., IsaacLab \citep{mittal2025isaaclab}, MuJoCo Playground \citep{zakka2025mujoco}, and HumanoidBench \citep{sferrazza2024humanoidbench}), these approaches successfully solve challenging locomotion and manipulation tasks that were once intractable. While recent works, such as FlashSAC \citep{kim2026flashsac}, have attempted to scale stochastic policy algorithms via system-level throughput optimizations, deterministic methods remain the dominant and most reliable baselines for very high-dimensional humanoid whole-body control.

Despite the dominance of deterministic methods, maximum entropy RL algorithms (e.g., SAC \citep{haarnoja2018soft}), theoretically offer compelling advantages, particularly in escaping local optima and fostering diverse behaviors, which are often absent in deterministic approaches.
However, realizing these benefits in high-dimensional humanoid control is limited by the ``curse of dimensionality'' \citep{koppen2000curse}.
A primary obstacle in complex robotic systems is that unconstrained exploration fails to distinguish between precision-critical and redundant actuators \citep{zhang2024on}.
Furthermore, because standard diagonal Gaussian policies lack structural prioritization, they distribute independent exploration noise uniformly across the expanded action volume ($|\mathcal{A}| > 20$).
As dimensionality increases, these independent perturbations severely restrict the robot's effective movement, termed ``vanishing exploration'' \citep{schumacher2023naturalrobustwalkingusing,wei2026scalableexplorationhighdimensionalcontinuous}.
Consequently, without a mechanism to selectively allocate exploration variance, stochastic agents frequently suffer from training instability and suboptimal convergence, a limitation shared by several recent strong baselines~\citep{sferrazza2024humanoidbench,seo2025fasttd3,kang2025entropyregularizingactivationboosting,wei2026scalableexplorationhighdimensionalcontinuous,kim2026flashsac}.

Compounding the exploration difficulty is the inherent susceptibility of critic networks to severe value overestimation in high-dimensional action spaces. As the action space expands, critics are frequently queried on state-action pairs lying far from the training distribution (OOD). In these sparse regions, neural approximations suffer from severe extrapolation error, generating spurious Q-value spikes \citep{Bhatt2019CrossQBN,neumann2024vlearnoffpolicylearning}. This issue is exacerbated in high-dimensional settings where the policy tends to over-exploit these approximation artifacts, aggressively shifting towards regions where the critic erroneously predicts high returns \citep{seo2025coarsetofineqnetworkactionsequence}. Crucially, standard mitigation strategies, such as clipped double Q-learning, are often insufficient to suppress this overestimation bias in complex dynamics \citep{kuznetsov2020controllingoverestimationbiastruncated,Duan2020DistributionalSA}.
Even recent parallelized approaches, utilizing either standard architectures or discrete distributional critics \citep{obandoceron2025simplicialembeddingsimprovesample,kang2025entropyregularizingactivationboosting}, struggle to resolve this instability, leading to suboptimal convergence.
Beyond overestimation, discrete critics rely on fixed supports and discrete atoms: when reward scales vary widely across tasks, the support range itself becomes an additional tuning burden, while quantization errors induced by finite atoms can further exacerbate value overestimation. Such discretization artifacts further compound extrapolation errors, compromising the accurate value estimation required for high-precision control. This motivates a continuous distributional critic, which can better preserve value fidelity while mitigating both high-dimensional overestimation and quantization artifacts in high-throughput settings \citep{DuanDSACv2}.

To bridge these gaps, we introduce FastDSAC, a framework that effectively scales maximum entropy stochastic policies to high-dimensional continuous control.
FastDSAC integrates two core mechanisms to unlock the scalability and stability of learning in high-dimensional regimes.
First, we propose Dimension-wise Entropy Modulation (DEM), which enables the agent to autonomously redistribute the exploration budget across action dimensions.
By effectively pruning the exploration subspace to suppress task-irrelevant noise, DEM avoids the structural inefficiency of uniform exploration typical of standard policies.
Second, we incorporate a continuous distributional critic parameterized by a Gaussian distribution \citep{DuanDSACv2}. 
Distinct from baselines limited to discrete approximations or deterministic estimates, this component ensures highly accurate value estimation by modeling the full return distribution continuously, avoiding fixed-support discretization while helping mitigate the value overestimation caused by high-dimensional extrapolation errors. 
Extensive empirical evaluations on HumanoidBench, MuJoCo Playground, and IsaacLab confirm that FastDSAC consistently matches or outperforms strong baselines across the vast majority of evaluated tasks. It yields performance gains of approximately \textbf{180\%} and \textbf{350\%} over FastTD3 on the challenging \textit{Basketball} and \textit{Balance Hard} tasks, respectively, demonstrating that rigorously designed stochastic policies can achieve outstanding performance in very high-dimensional humanoid robotics. Furthermore, we validate the practical viability of our framework by successfully deploying the learned policies zero-shot to a physical Unitree G1 humanoid. An anonymous implementation of our method is provided at \url{https://anonymous.4open.science/r/FastDSAC_official-C43F}.

\section{Preliminaries}
\label{sec:preliminaries}

We formulate the continuous control problem as a Markov Decision Process (MDP), defined by the tuple $(\mathcal{S}, \mathcal{A}, p, r, \gamma)$, comprising the state space $\mathcal{S}$, action space $\mathcal{A}$, transition dynamics $p(s'|s, a)$, reward function $r(s, a)$, and discount factor $\gamma \in [0, 1)$.
Our approach integrates two RL paradigms:

\textbf{Maximum Entropy Reinforcement Learning.} To encourage exploration, we adopt the maximum entropy framework \citep{haarnoja2018soft}. Instead of solely maximizing scalar returns, the agent learns a policy $\pi(\cdot|s)$ optimizing the expected return augmented by an entropy term:
\begin{equation}
    J(\pi) = \mathbb{E}_{\pi} \left[ \sum_{t=0}^\infty \gamma^t \Big( r(s_t, a_t) + \alpha \mathcal{H}(\pi(\cdot|s_t)) \Big) \right],
\end{equation}
where $\mathcal{H}$ is the policy entropy and $\alpha$ is a temperature parameter balancing reward and exploration.

\textbf{Distributional Reinforcement Learning.} Unlike standard methods estimating the expected value $Q^\pi(s, a) = \mathbb{E}[Z^\pi(s, a)]$, we model the full distribution of the random return $Z^\pi(s, a)$, which follows the distributional Bellman equation:
\begin{equation}
    Z^\pi(s, a) \overset{D}{=} r(s, a) + \gamma Z^\pi(s', a'),
\end{equation}
where $s' \sim p(\cdot|s, a)$, $a' \sim \pi(\cdot|s')$, and $\overset{D}{=}$ denotes distributional equality. Capturing return distribution enables the agent to utilize higher-order moments (e.g., variance) for more stable learning signals.

\subsection{Fast Actor-Critic (FastTD3)}
Our work builds on the high-throughput recipe of FastTD3~\citep{seo2025fasttd3}, a TD3 variant designed for massive parallelism. Two ingredients are especially relevant for FastDSAC. (1) FastTD3 leverages massively parallel simulation together with large-batch replay updates, which accelerate training and increase data diversity within each critic update. (2) It equips TD3 with the categorical C51 critic~\citep{bellemare2017distributionalperspectivereinforcementlearning}, which represents returns on a fixed discrete support of $N$ atoms $\{z_i\}_{i=1}^{N}$ spanning $[v_{\min}, v_{\max}]$. Target estimation uses the projected Bellman update $\boldsymbol{m} = \Phi \mathcal{T} Z_{\theta'}$, where the distributional Bellman operator $\mathcal{T}$ first shifts the return distribution and $\Phi$ then projects it back onto the fixed atoms; the full C51 parameterization and optimization objective are summarized in Appendix~\ref{app:fasttd3_dsact_background}.

However, this fixed discretization is limiting in our setting: the support range $[v_{\min}, v_{\max}]$ can become task-dependent when reward scales vary, and finite atoms introduce quantization error that restricts value estimation precision in continuous control tasks requiring fine-grained signal propagation.
\subsection{Continuous Distributional Critics (DSAC)}
To mitigate the limitations of discrete distributional approximations and ensure accurate value estimation, we employ the Distributional Soft Actor-Critic (DSAC) framework \citep{Duan2020DistributionalSA,DuanDSACv2}. DSAC parameterizes the return distribution as a continuous Gaussian $Z_\theta(\cdot|s, a) \sim \mathcal{N}(Q_\theta(s, a), \sigma_\theta^2(s, a))$, thereby avoiding both quantization artifacts and manual support specification. 
Furthermore, to address the numerical instability of standard KL-divergence gradients, DSAC derives separate gradient estimators for the mean and standard deviation. This separation inversely scales the mean update by variance to reduce step sizes in high-uncertainty regions, thereby curbing overestimation, while stabilizing the variance update via target clipping $C(\cdot; b)$ to prevent optimization collapse \citep{DuanDSACv2}: 
\begin{equation}
\label{eq:dsact_grad}
    \nabla_{\theta} J_Z \approx (\omega + \epsilon_\omega) \mathbb{E} \biggl[ \underbrace{- \frac{y_q^{\min} - Q_{\theta}}{\sigma_{\theta}^2 + \epsilon} \nabla_{\theta} Q_{\theta}}_{\text{(I) Mean Update with Expected Value}} 
     \underbrace{- \frac{(C(y_z^{\min}; b) - Q_{\theta})^2 - \sigma_{\theta}^2}{\sigma_{\theta}^3 + \epsilon} \nabla_{\theta} \sigma_{\theta}}_{\text{(II) Variance Update with Clipping}} \biggr].
\end{equation}
Here, $y_q^{\min}$ denotes the conservative target Q-value derived from the twin critics, while $y_z^{\min}$ represents a stochastic sample from the target return distribution. The parameter $\epsilon$ prevents gradient explosion when the variance $\sigma_{\theta}^2$ approaches zero, whereas $\epsilon_\omega$ prevents gradient disappearance as the scaling factor $\omega$ approaches zero. The scaling factor $\omega=\mathbb{E}_{(s,a)\sim\mathcal{B}}[\sigma_\theta(s,a)^2]$ and the clipping boundary $b$ within $C(\cdot; b)$ are adaptively updated based on the running statistics of the learned variance. As detailed in DSAC-T~\citep{DuanDSACv2}, this estimator combines three mechanisms: Expected Value Substituting, Twin Value Distribution Learning, and Variance-Based Critic Gradient Adjustment. Appendix~\ref{app:fasttd3_dsact_background} summarizes their original formulation, while Section~\ref{sec:method_critic} explains how FastDSAC streamlines them for the high-throughput regime.

\section{Methodology}
\label{sec:method}

In this section, we present FastDSAC, a high-throughput reinforcement learning framework tailored for high-dimensional continuous control. FastDSAC adapts the actor-critic paradigm to massive parallelization via structural innovations in both policy parameterization and value estimation. The overall architecture diagram of the algorithm and the pseudocode are shown in Appendix~\ref{app:algorithm}. 

Our method consists of three integrated components: (1) An actor employing Dimension-wise Entropy Modulation to autonomously manage exploration budgets in high-dimensional action spaces (Section~\ref{sec:method_actor}); (2) A streamlined Continuous Distributional Critic that improves value
fidelity by avoiding fixed-support discretization and reducing quantization artifacts (Section~\ref{sec:method_critic}); and (3) A unified Distributional Soft Policy Iteration (DSPI) loop optimized for large-batch stability (Section~\ref{sec:method_dspi}).

\subsection{The FastDSAC Actor: Dimension-wise Entropy Modulation}
\label{sec:method_actor}  
Standard Gaussian policies in SAC typically parameterize the action distribution as $\pi_\phi(a|s) = \mathcal{N}(\mu_\phi(s), \text{diag}(\sigma_\phi^2(s)))$, estimating the standard deviation $\sigma_{\phi,i}$ independently for each dimension. In high-dimensional systems, this independent parameterization leads to inefficient exploration, as the agent fails to distinguish between critical and redundant actuators, assigning variance uniformly across all dimensions \citep{zhang2024on,sferrazza2024humanoidbench}.
To address this, we propose Dimension-wise Entropy Modulation (DEM), a mechanism that imposes a structural constraint on the exploration covariance to explicitly manage the variance budget. 
Rather than replacing the standard variance prediction, DEM augments it by learning a dimension-specific scaling weight $w_i$ for each action dimension $i \in \{1, \dots, |\mathcal{A}|\}$.

\textbf{Budget-Preserving Variance Redistribution.} Given the actor backbone features $h(s)$, we compute the modulation weights using a temperature-scaled Softmax operation:
\begin{equation} \label{eq:weights}
l_i = f_\phi^{\mathrm{logits}}(h(s))_i, \quad w_i = \frac{\exp\left(l_i / \tau\right)}{\sum_{j=1}^{N} \exp\left(l_j / \tau\right)} \cdot N,
\end{equation}
where $\tau$ is a temperature hyperparameter that regulates the sparsity of the weight distribution, and $N = |\mathcal{A}|$ denotes the action dimension.
Multiplying the standard Softmax output by $N$ ensures that the mean of the weights is exactly $1$ (i.e., $\frac{1}{N} \sum_{i=1}^{N} w_i = 1$). This normalization enforces a zero-sum redistribution of the exploration budget: increasing the variance on specific redundant dimensions necessarily decreases the variance on precision-critical dimensions.
The final standard deviation for the $i$-th action dimension is modulated as:
\begin{equation} \sigma_{\phi,i}(s) = w_i(s) \cdot \exp\left( \hat{\sigma}_{\phi,i}(s) \right), \end{equation}
where $\hat{\sigma}_{\phi,i}(s)$ represents the base log-standard deviation predicted by the network. The subsequent action sampling and policy evaluation follow the standard SAC reparameterization trick.

\textbf{Interpretation: Autonomous Subspace Pruning.} This formulation enforces a soft exploration budget.
By dynamically adjusting $w_i$, the agent gains the autonomy to selectively suppress noise on specific dimensions while concentrating exploration variance on task-irrelevant dimensions, effectively rendering critical actions near-deterministic to ensure the precise control for optimal convergence.
Crucially, this importance assignment is emergent and driven solely by reward maximization. This bypasses the need for manual priors or kinematic specifications, enabling the discovery of effective coordination patterns that may be non-intuitive to human design.
As empirically analyzed in Section~\ref{sec:ablation}, this mechanism effectively prunes the exploration subspace, maintaining high-precision control on stable joints while exploring others.
The temperature $\tau$ regulates the \textit{sparsity} of this weight distribution: a lower $\tau$ encourages a sharp, focused allocation, while a higher $\tau$ yields a uniform distribution, approximating standard SAC behavior.

\textbf{Diversity-driven Population.} To mitigate mode collapse across the parallel population, we employ a Heterogeneous Exploration scheme adapted from FastTD3~\citep{seo2025fasttd3}. We assign a distinct scaling factor $\beta_e \sim U[\beta_{\min}, \beta_{\max}]$ to each environment $e$. 
 Unlike prior works that scale noise magnitude \citep{li2023parallel,seo2025fasttd3}, we apply $\beta_e$ to the modulation logits to vary the distribution shape:
\begin{equation} 
w_{i,e} \propto \exp\left( \frac{l_i \cdot \beta_e}{\tau} \right). 
\end{equation}
This effectively diversifies the anisotropy of exploration across the population: some agents explore broadly by distributing the budget across many dimensions, while others focus intensely on specific subspaces, thereby enhancing robustness against local optima.

\subsection{The FastDSAC Critic: Streamlined Distributional Learning}
\label{sec:method_critic}  
To address the limitations of discrete critics \citep{bellemare2017distributionalperspectivereinforcementlearning,seo2025fasttd3}, we adopt the Continuous Gaussian Critic from DSAC-T \citep{DuanDSACv2}, parameterizing the return distribution as $Z_\theta(\cdot|s,a) \sim \mathcal{N}(Q_\theta(s,a), \sigma_\theta^2(s,a))$, to avoid fixed-support specification and ensure accurate value estimation by modeling the full return distribution without discretization artifacts.

While original DSAC-T employs complex variance-based gradient adjustments for low-data regimes, we find these constraints restrict accurate variance modeling in our settings.
Leveraging the natural stability provided by large batch sizes, we streamline the objective by removing variance clipping boundaries, while retaining twin-critic target, Expected Value Substitution, and Gradient Scaling.

The parameters $\theta$ are updated by minimizing divergence between predicted and target distributions. Following DSAC-T~\citep{DuanDSACv2}, we decompose the gradient into mean- and variance-driven components:
\begin{equation}
\label{eq:fastdsac_critic_grad}
\nabla_{\theta} J_{Z} \approx \omega \cdot \mathbb{E} \bigg[ \bigg( \underbrace{ - \frac{y_q^{\min} - Q_\theta}{\sigma_\theta^2 + \epsilon} \nabla_{\theta} Q_\theta }_{\text{(I) Stabilized Mean Update}} 
 \underbrace{ - \frac{(y_z^{\text{sample}} - Q_\theta)^2 - \sigma_\theta^2}{\sigma_\theta^3 + \epsilon} \nabla_{\theta} \sigma_\theta }_{\text{(II) Natural Variance Update}} \bigg) \bigg].
\end{equation}
Here, term (I) leverages the Expected Value Substitution: we anchor the mean update to the conservative target mean $y_q^{\min} = r + \gamma (\min_{j=1,2} Q_{\theta'j}(s', a') - \alpha \log \pi(a'|s'))$, rather than stochastic samples. In our high-dimensional setting, this deterministic target proved more stable than directly using stochastic return samples for the mean update.
Term (II) updates the variance to capture the stochasticity induced by environmental dynamics and policy exploration.
Unlike the original implementation which clips the target sample $y_z^{\text{sample}}$ within a dynamic boundary, we remove this constraint to allow the critic to learn the true scale of environmental stochasticity without bias.
The scalar $\omega \approx \mathbb{E}_{\mathcal{B}}[\sigma_\theta^2]$ is the gradient scaling factor handling reward scale sensitivity \citep{DuanDSACv2}.

\subsection{Distributional Soft Policy Iteration (DSPI)}
\label{sec:method_dspi}
We integrate the proposed Actor~\ref{sec:method_actor} and Critic~\ref{sec:method_critic} into a unified Distributional Soft Policy Iteration (DSPI) loop.
This process alternates between evaluating the policy's return distribution and improving the policy via entropy-regularized maximization. Theoretically, the DSPI framework is guaranteed to converge to the optimal policy, as established in DSAC~\citep{Duan2020DistributionalSA}.

\textbf{Distributional Soft Policy Evaluation.} In the evaluation step, we update the twin critic networks using the streamlined gradient estimator in Eq.~\ref{eq:fastdsac_critic_grad}. 
By maintaining a continuous Gaussian parameterization, the critic avoids discrete quantization artifacts and models the full return distribution, capturing aleatoric uncertainty alongside the expected return. As the gradient magnitude is inversely scaled by the estimated variance, high uncertainty on OOD actions naturally dampens updates, reducing high-dimensional overestimation and improving value estimation.

\textbf{Distributional Soft Policy Improvement.}  In the improvement step, the actor $\pi_\phi$ is updated to maximize both the expected return and entropy.
Following the reparameterization trick \citep{haarnoja2018soft}, the actor utilizes the mean estimate of the distributional critic, $Q_\theta(s,a) = \mathbb{E}_{Z \sim Z_\theta}[Z]$, to guide policy updates.
The actor is optimized by minimizing:
\begin{equation}
\mathcal{L}_{\text{actor}}(\phi) = \mathbb{E}_{s \sim \mathcal{B}, a \sim \pi_\phi} \left[ \alpha \log \pi_\phi(a|s) - Q_\theta(s,a) \right].
\end{equation}
Since our actor employs the Dimension-wise Entropy Modulation (DEM) mechanism, the term $\log \pi_\phi(a|s)$ implicitly accounts for the redistributed exploration budget. This optimization naturally balances reward maximization against the structural entropy constraints imposed by DEM, effectively guiding the agent to autonomously prune task-irrelevant subspaces.
Simultaneously, we dynamically adjust the temperature parameter $\alpha$ via gradient descent to satisfy the target entropy constraint $\mathcal{H}$ \citep{haarnoja2018soft}:
\begin{equation}
\alpha \leftarrow \alpha - \eta \cdot \mathbb{E}_{s \sim \mathcal{B}, a \sim \pi_\phi} \left[ -\log \pi_\phi(a|s) - \mathcal{H} \right]
\end{equation}
\textbf{Implementation Recipe Summary.} Full implementation details are provided in Appendix~\ref{app:implementation_details}. Crucially, we set the target entropy to $\mathcal{H} = 0$ across all tasks. Unlike the standard heuristic ($\mathcal{H} = -|\mathcal{A}|$) which forces premature variance decay, $\mathcal{H} = 0$ sustains a generous exploration budget, which DEM structurally manages by safely offloading excess entropy to redundant joints. Additionally, we apply LayerNorm only in the very high-dimensional HumanoidBench domain, where it serves as an important stabilizer for training, whereas smaller action spaces (MuJoCo Playground \& IsaacLab) achieve SOTA performance without such regularization. Furthermore, unlike the original DSAC-T \citep{DuanDSACv2}, which requires a large numerical stabilizer (bias $= 0.1$) to prevent division-by-zero, our large-batch regime allows drastically reducing the continuous critic's numerical stabilizer (bias $= 10^{-6}$), unlocking high-fidelity value estimation for low-variance states. Finally, while a fixed DEM temperature ($\tau=1$) is our robust default, we introduce an end-to-end learnable variant, FastDSAC (auto-$\tau$), to adapt variance redistribution for highly dynamic tasks. Ablation experiments on target entropy and layernorm are in Appendix~\ref{sec:appendix_layernorm_ablation} and ~\ref{sec:appendix_target_entropy_ablation}.

\section{Experiments}
\label{sec:experiments}

Our empirical evaluation aims to validate the scalability, algorithmic mechanisms, and real-world viability of FastDSAC in high-dimensional continuous control. Specifically, we design our experiments to address four key questions:
\textbf{Q1 (Comparative Performance):} Can FastDSAC match or outperform strong deterministic, standard stochastic, and model-based baselines across complex domains?
\textbf{Q2 (Actor Mechanism):} Is Dimension-wise Entropy Modulation (DEM) essential for managing high-dimensional exploration, and does it autonomously discover interpretable variance-pruning strategies?
\textbf{Q3 (Critic Parameterization):} Does continuous Gaussian modeling yield more accurate value estimation in high-dimensional spaces compared to discrete (C51) approximations?
\textbf{Q4 (Sim-to-Real Deployment):} Can the stable, high-precision control policies discovered by FastDSAC in simulation transfer zero-shot to physical humanoid hardware?

To address these questions, we evaluate our framework across HumanoidBench, MuJoCo Playground, and IsaacLab, spanning diverse tasks from locomotion to complex whole-body manipulation, and ultimately deploy it on the Unitree G1.

\subsection{Experimental Setup}

\textbf{Benchmarks.} We aligned with the FastTD3 suite \citep{seo2025fasttd3}, evaluating on:
(1) HumanoidBench \citep{sferrazza2024humanoidbench}: 25 representative tasks spanning locomotion and whole-body manipulation, with very high-dimensional action spaces ($|\mathcal{A}| = 61$).
(2) MuJoCo Playground \citep{zakka2025mujoco} \& IsaacLab \citep{mittal2025isaaclab}: 10 tasks across diverse physics engines to verify robustness. While their action spaces are smaller than those of HumanoidBench ($|\mathcal{A}| < 30$), they still represent challenging high-dimensional control settings in RL. We therefore omitted Layer Normalization, as FastDSAC retains inherent stability without it.

\textbf{Baselines \& Protocol.} We benchmarked against FastTD3 (deterministic SOTA) and FastSAC (parallelized stochastic baseline) \citep{seo2025fasttd3}. To ensure rigorous comparison in all tasks, we matched FastSAC to FastDSAC in its Layer Normalization setting and target entropy $\mathcal{H}=0$, hereafter denoted as \textit{FastSAC (standard)}.
For HumanoidBench, we incorporated official benchmark results for DreamerV3, TD-MPC2, SAC, and PPO; for MuJoCo Playground and IsaacLab, we additionally evaluated against PPO to contextualize performance against on-policy methods in these domains.

To guarantee fair evaluation, we strictly inherited the FastTD3 high-throughput training protocol (e.g., replay capacity, update schedules, parallel environment counts) for all corresponding tasks. We report mean returns aggregated over 3 seeds, with shaded regions indicating the min-max range. Notably, we use 5 seeds for \textit{Basketball} and \textit{Balance Hard} to demonstrate consistency, while maintaining 3 seeds for all other tasks to match FastTD3 and HumanoidBench protocols. Crucially, for FastDSAC-specific hyperparameters, we rigorously enforce \textbf{unified algorithmic configuration} within each evaluation domain (i.e., parameters remain strictly identical across all tasks within a given benchmark), in contrast to FastTD3, whose distributional critic may require task-specific tuning of the C51 support range. Full protocol details and hyperparameter configurations are summarized in Appendix~\ref{app:hyperparameters}.

\subsection{Main Results (Q1)}
\label{sec:main_results}
\begin{figure}[t]
    \centering
    \includegraphics[width=1.0\textwidth]{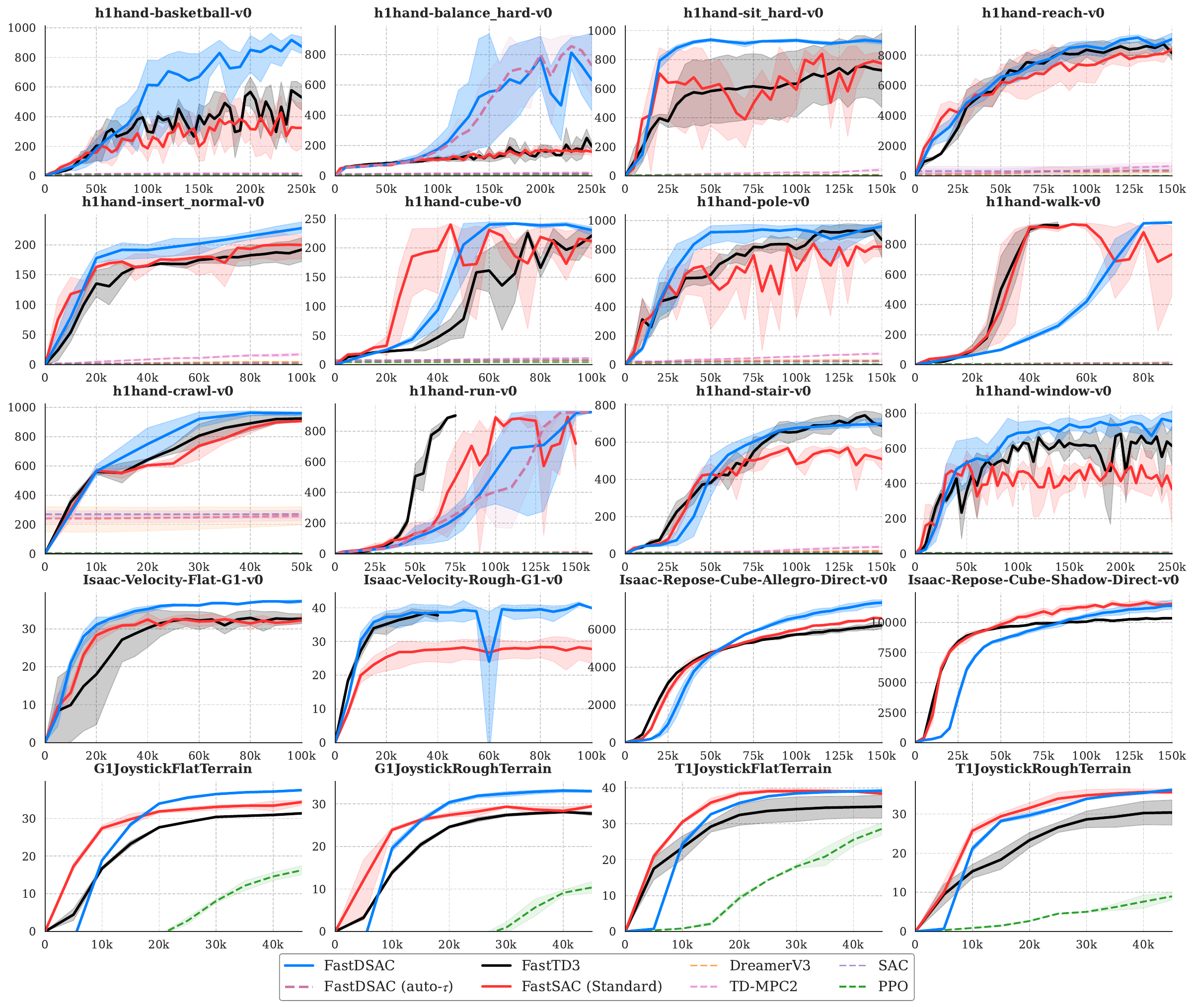} 
    \caption{\textbf{Comparative evaluation on high-dimensional continuous control benchmarks.} 
    Learning curves across selected tasks from \textbf{HumanoidBench} (top three rows), \textbf{IsaacLab} (fourth row), and \textbf{MuJoCo Playground} (bottom row). 
    \textbf{FastDSAC} matches or often outperforms \textbf{FastTD3} and \textbf{FastSAC (Standard)}. It performs strongly on precision-demanding (e.g., \textit{Basketball}, \textit{Insert}) and stability-critical (e.g., \textit{Balance Hard}) tasks, while retaining robust dexterous manipulation and rough-terrain locomotion performance.}
    \label{fig:main_results}
\end{figure}

We evaluated FastDSAC across 35 tasks spanning diverse dynamics. Figure~\ref{fig:main_results} shows learning curves for 20 representative tasks from HumanoidBench, MuJoCo Playground, and IsaacLab (see Appendix~\ref{app:full_results} for full results). We observe three distinct performance regimes:

\textbf{Strong Performance on Complex Coordination \& Manipulation.} 
FastDSAC achieves state-of-the-art performance on tasks demanding precise actuation and whole-body coordination, outperforming FastTD3 and FastSAC (Standard) on challenging tasks such as \textit{Basketball}, \textit{Balance Hard}, \textit{Insert Normal}, \textit{Sit Hard}, and \textit{Window}.
This gap highlights the limitation of deterministic policies in exploring intricate control manifolds, whereas FastDSAC's entropy modulation effectively prunes the exploration subspace to discover superior solutions that deterministic baselines often fail to reach.

\textbf{The Exploration-Exploitation Trade-off in Locomotion.} 
In dynamic locomotion (e.g., \textit{Run}, \textit{slide}), FastDSAC rises more slowly than FastTD3 but converges to higher final returns.
This pattern is consistent with the maximum entropy objective: DEM initially allocates variance to explore diverse gait patterns before stabilizing, whereas FastTD3 often converges prematurely to suboptimal gaits.
Even on complex whole-body control tasks such as \textit{Stair}, FastDSAC remains competitive with FastTD3, showing that our framework preserves the precision needed for rigorous terrain adaptation.

\textbf{Robustness Across Simulators.} 
FastDSAC scales consistently across physics engines.
In MuJoCo Playground (e.g., \textit{G1/T1 Joystick}), it maintains a clear lead. 
In IsaacLab (\textit{Velocity-Rough}, \textit{Repose-Cube-Allegro/Shadow}), FastDSAC outperforms FastTD3 after an initial adaptation phase, validating its robustness across dynamics.
Although FastSAC (Standard) converges faster on the lower-dimensional \textit{Repose-Cube-Shadow} task ($|\mathcal{A}| \approx 24$), FastDSAC reaches comparable performance, suggesting that while unmodulated exploration can suffice for specific hand-only tasks, our framework remains the stronger generalist across the broader benchmark suite.

\subsection{Ablation Studies and Analysis (Q2 \& Q3)}
\label{sec:ablation}
\begin{wrapfigure}[25]{r}{0.5\textwidth}\centering\includegraphics[width=\linewidth]{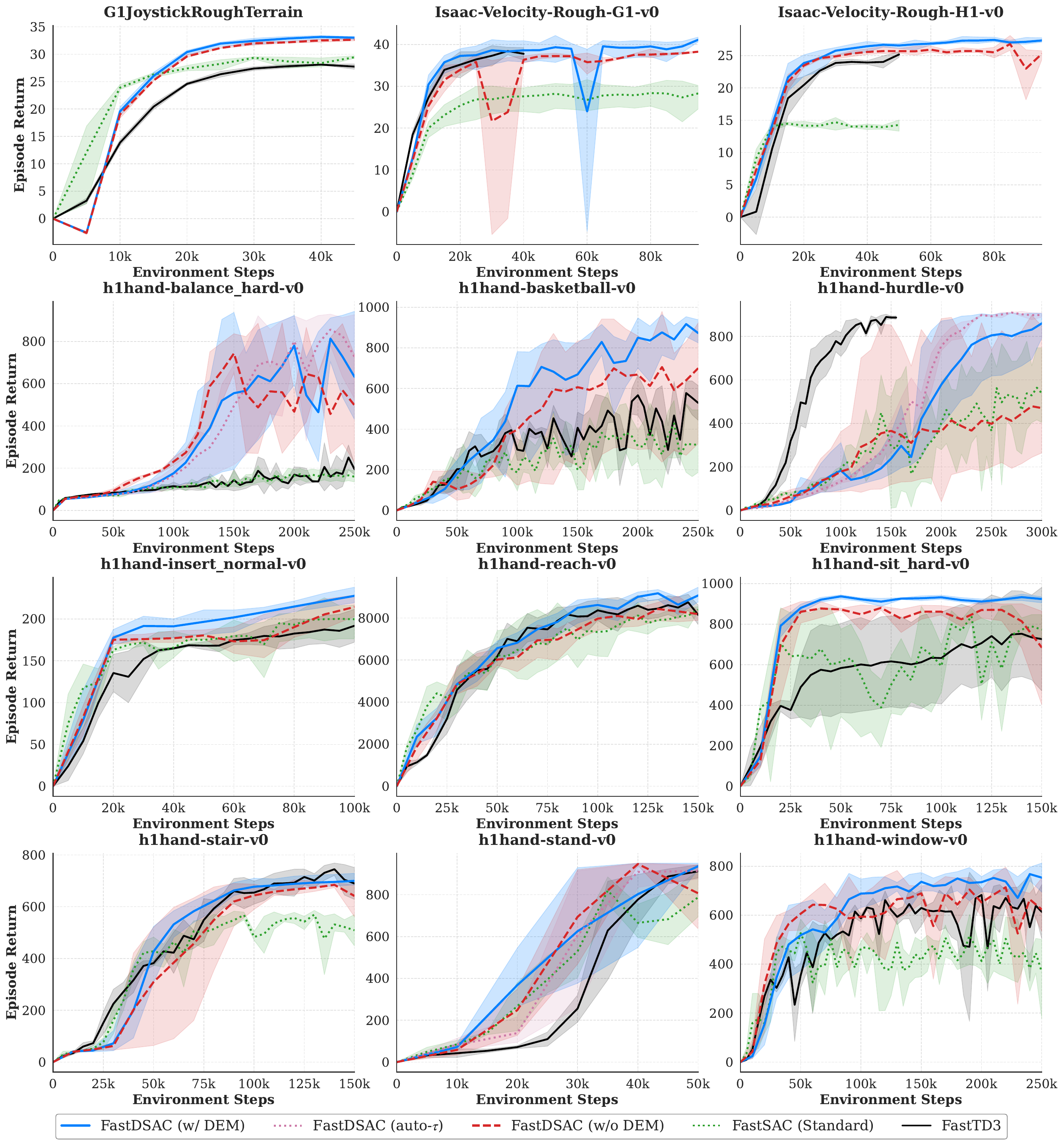}\caption{\textbf{DEM improves performance and stability in high-dimensional control.} Learning curves across 12 tasks from three benchmarks comparing FastDSAC, its ablations (w/o DEM and auto-$\tau$), and FastTD3/FastSAC.}\label{fig:actor_ablation}
\end{wrapfigure}
To assess component necessity, we conducted ablation studies across 3 benchmarks on representative tasks spanning rough-terrain locomotion, dynamic whole-body coordination, object interaction, and manipulation.

\textbf{Q2: Necessity and Mechanism of Dimension-wise Entropy Modulation (DEM).} 
\label{sec:ablation:actor_ablation}
To isolate the effect of DEM, we compare FastDSAC against an ablated variant (w/o DEM) that replaces the entropy-modulated actor with a standard diagonal Gaussian policy. As shown in Figure~\ref{fig:actor_ablation}, across 12 diverse tasks from all 3 benchmarks, the full FastDSAC consistently yields higher asymptotic returns. Although the ablated agent often outperforms standard baselines, likely due to the continuous distributional critic, removing DEM degrades performance, reduces training stability, and increases inter-seed variance in tasks with challenging dynamics (e.g., \textit{Balance Hard}, \textit{Basketball}, and \textit{Hurdle}). Beyond stabilizing the baseline performance, the structure of DEM also supports end-to-end optimization of its temperature parameter, which can further improve returns on tasks with different exploration demands. These results show DEM is indispensable for managing high-dimensional exploration and reducing seed sensitivity.

\textbf{Visualizing the Mechanism: Autonomous Subspace Pruning.}
To understand how DEM optimizes exploration, we analyzed the evolution of learned weights $w_i$ during a \textit{Basketball} episode (see the qualitative comparison in Figure~\ref{fig:comp_basketball} and the heatmap in Appendix~\ref{app:basketball_episode_heatmap}).
The visualization reveals a task-aligned exploration strategy under the budget-preserving constraint ($\frac{1}{N}\sum w_i = 1$), which enforces a zero-sum redistribution: to achieve near-deterministic precision on critical joints, the agent must offload variance onto redundant dimensions.
\textbf{1) Variance Offloading:} We observed a prominent high-weight band (reaching $w_i \approx 8.0$) corresponding to task-irrelevant joints, such as the left thumb (\texttt{lh\_A\_THJ4}) and wrist, which act as entropy sinks. 
\textbf{2) Precision for Impact and Balance:} During the critical interaction phase (Steps 10-13), as the robot impacts the ball, weights for the legs, torso, and right arm are suppressed to near-zero. This suppression persists post-impact to counteract momentum, demonstrating DEM's ability to autonomously prune the active subspace to guarantee whole-body stability.
\begin{wrapfigure}[12]{r}{0.35\textwidth}\centering\includegraphics[width=\linewidth]{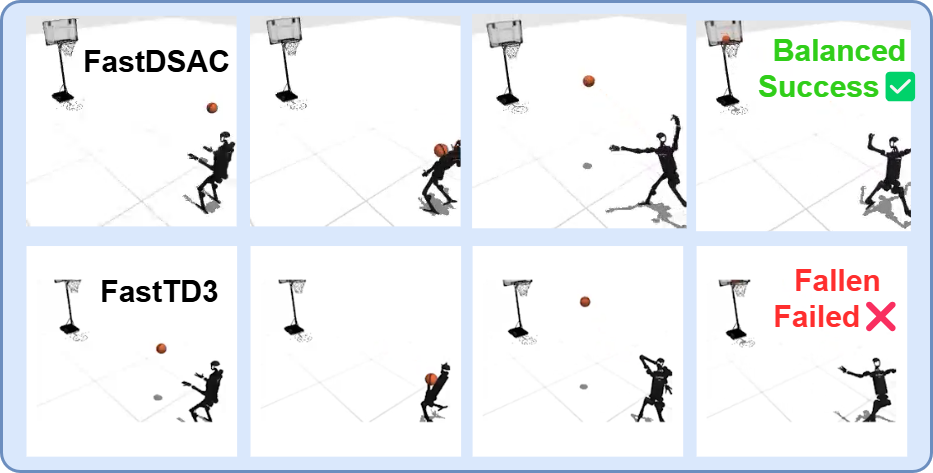}\caption{\textbf{Qualitative comparison on \textit{Basketball}.} FastDSAC throws well and stays stable, while FastTD3 loses balance and fails.}
\label{fig:comp_basketball}
\end{wrapfigure}

\textbf{Interpretation: Emergent Strategy via Variance Offloading.} 
The agent autonomously discovers an unconventional ``body-rebound'' strategy, using the torso rather than the hands to redirect the ball. While counter-intuitive to human design, this behavior maximizes return by prioritizing post-throw stability over risky dexterity.
The heatmap shows that the agent treats the task-irrelevant left thumb as an \textit{entropy sink} for the required exploration budget.
By offloading variance to these distal joints, FastDSAC effectively prunes the exploration subspace, preserving high-precision control for the core body and legs to attain consistent returns of $\approx 1030$.
Crucially, despite kinematic variations across seeds, the variance redistribution pattern remains stable: FastDSAC repeatedly uses task-irrelevant subspaces as entropy sinks to stabilize whole-body control.

\textbf{Contrast with Deterministic Baseline.}
FastTD3 attempts an intuitive but unstable hand-catch strategy (Figure~\ref{fig:comp_basketball}). However, it fails to balance manipulation with whole-body stability; the agent hits the ball but then falls, resulting in suboptimal returns ($\approx 500$).
In contrast, FastDSAC's entropy modulation exploits the reward structure more effectively and finds a more stable solution. More broadly, this result shows that in high-dimensional RL, the return-maximizing strategy need not align with human-intended behavior, and may instead favor whole-body survival over risky dexterity.


\begin{wrapfigure}[20]{r}{0.45\textwidth}\centering\includegraphics[width=\linewidth]{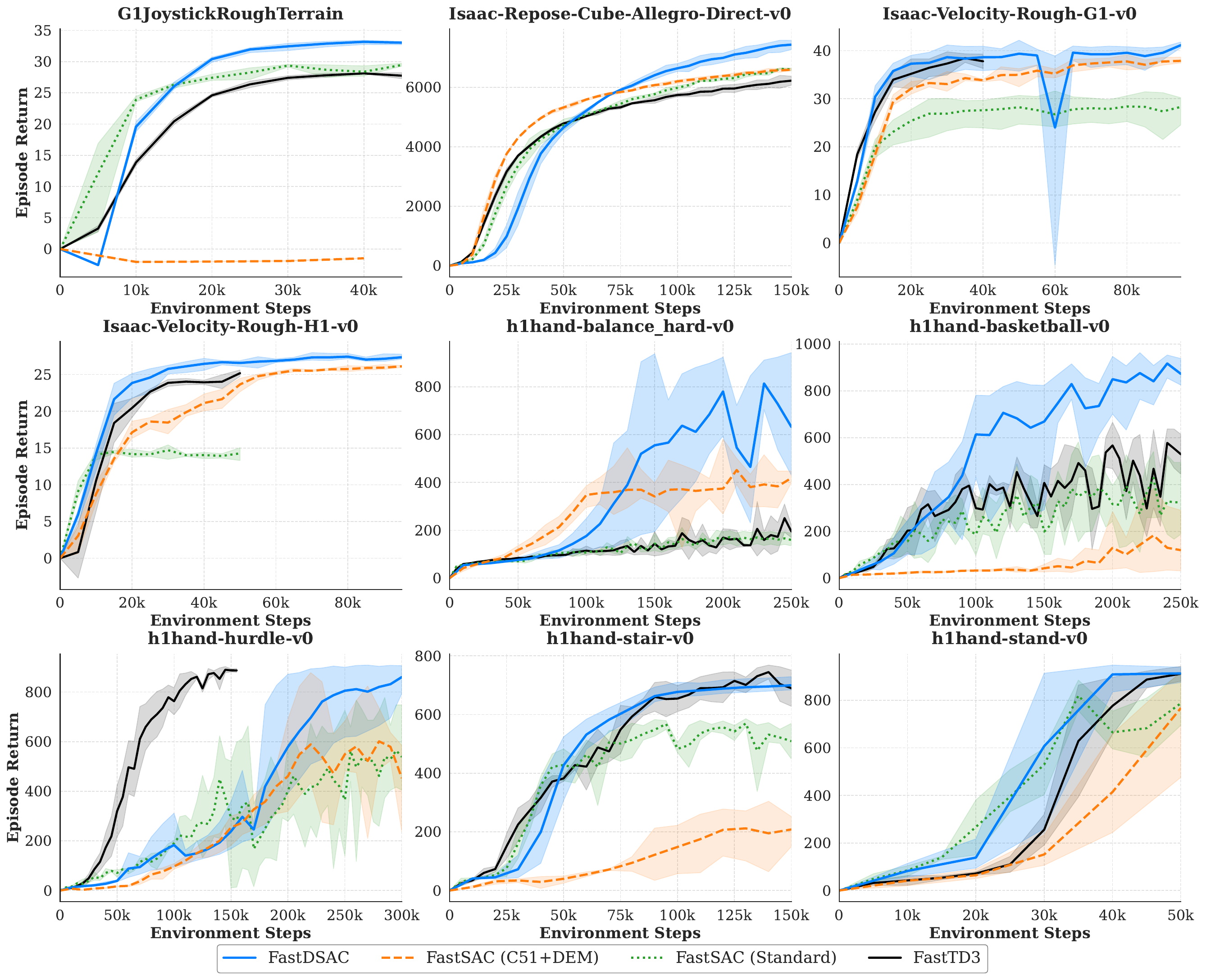}\caption{\textbf{Continuous critics outperform discrete C51 critics on 9 high-dimensional tasks.} Curves show the mean episode return of FastDSAC compared to FastSAC (C51$+$DEM) and FastSAC (standard).}
\label{fig:continuous}
\end{wrapfigure}
\textbf{Q3: Continuous vs. Discrete Distributional Critics.}  

To isolate the effect of critic parameterization, we compared FastDSAC against a discrete baseline, FastSAC (C51$+$DEM), the C51 variant of FastSAC~\citep{seo2025learningsimtorealhumanoidlocomotion} augmented with our DEM module, using the same training pipeline and actor design.
Figure~\ref{fig:continuous} shows that FastDSAC outperforms both FastSAC (standard) and its discrete variant, FastSAC (C51$+$DEM), on all nine evaluated tasks, including approximately $2\times$ higher returns on \textit{Balance Hard}. Since the actor design and the overall training framework are otherwise matched, these results suggest that the continuous Gaussian critic provides a more accurate and stable value estimate than the discrete C51 critic in our high-throughput control regime.
A likely reason is that the continuous critic avoids fixed-support discretization and atom quantization, which may introduce representation errors that degrade value accuracy in high-dimensional control. Consistent with this view, Appendix~\ref{app:c51_support} shows on \textit{Basketball} that the performance of C51$+$DEM is highly sensitive to the choice of support range. Moreover, prior DSAC analyses~\citep{DuanDSACv2} show that distributional critics can reduce value overestimation, which becomes particularly important in high-dimensional action spaces.


\begin{wrapfigure}[11]{r}{0.33\textwidth}\centering\vspace{-20pt}\includegraphics[width=\linewidth]{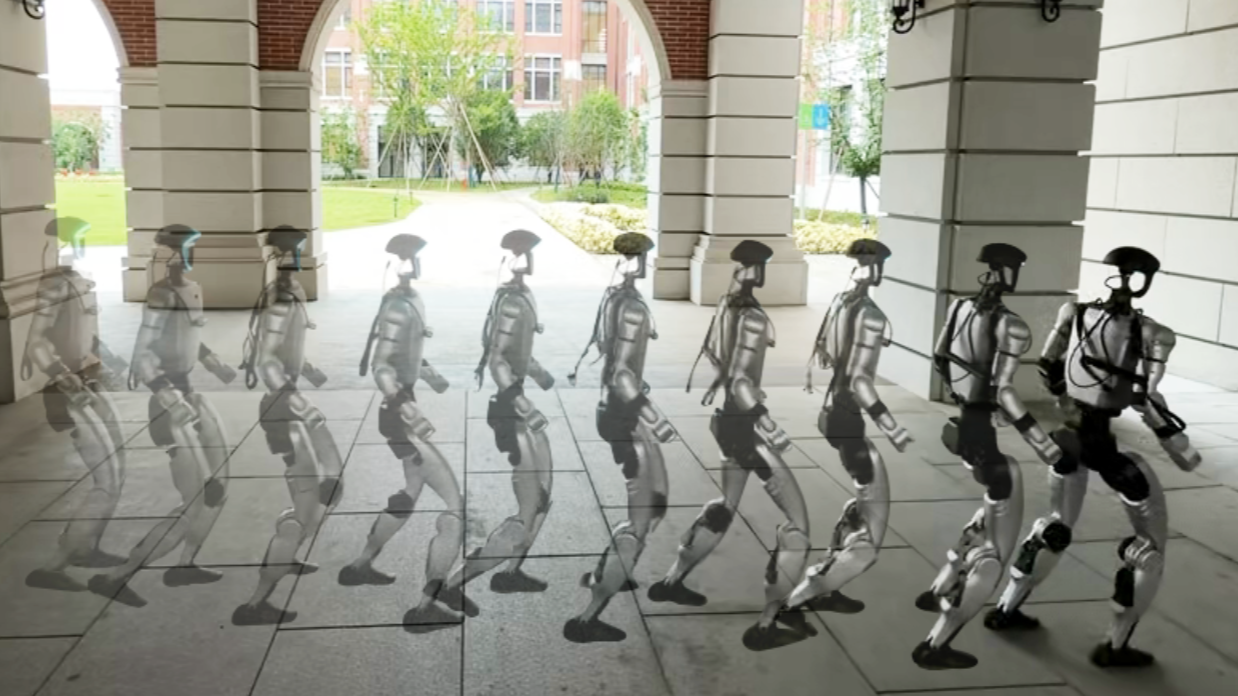}\caption{\textbf{Sim-to-real on Unitree G1.} Zero-shot joystick-conditioned forward walking on the G1 robot, showing effectiveness of FastDSAC in real-world deployments.}
\label{fig:g1_sim2real}
\end{wrapfigure}
\subsection{Sim-to-Real Validation on Unitree G1 (Q4)}
\label{sec:sim2real}
To verify that FastDSAC transfers effectively to physical hardware, we deploy FastDSAC zero-shot on a Unitree G1 humanoid, following the mature sim-to-real pipeline in \citep{seo2025learningsimtorealhumanoidlocomotion}. Evaluated on joystick-conditioned locomotion and a complex ``squat-carry-walk-squat'' motion-tracking task, FastDSAC requires minimal adaptation: core hyperparameters remain unified, adjusting only task-specific settings (e.g., multi-step returns, $\gamma$).
Running directly on the G1's onboard compute, the policies show zero-shot stability and robustness to moderate physical pushes (Figure~\ref{fig:g1_sim2real}). This shows that FastDSAC's stable exploration in simulation can yield deployable real-world controllers. See Appendix~\ref{app:sim2real} for motion-tracking and additional locomotion results.

\section{Conclusion}
\label{sec:conclusion}
In this work, we challenge the dominance of deterministic policy gradients in high-dimensional humanoid control by introducing FastDSAC, a framework that unlocks the scalability of maximum entropy RL. FastDSAC addresses the curse of dimensionality by integrating Dimension-wise Entropy Modulation (DEM) for autonomous subspace pruning, together with a continuous distributional critic for accurate value estimation by mitigating both high-dimensional overestimation and discretization artifacts. Across 35 diverse tasks, our method establishes state-of-the-art performance for high-dimensional stochastic policies on the evaluated benchmarks, remaining competitive with and often outperforming strong baselines, with gains of 180\% and 350\% on challenging coordination tasks. Furthermore, FastDSAC policies successfully transfer zero-shot to a physical humanoid robot. Our findings show that a rigorously designed stochastic policy is not only viable for the high-throughput regime, but can also achieve strong performance in high-precision humanoid control, bridging the gap between broad exploration and precise control. Detailed discussion of limitations, extended future directions, and broader societal impacts are provided in Appendix~\ref{sec:limitations}. 


\bibliography{reference}
\bibliographystyle{unsrtnat}



\newpage
\thispagestyle{plain}
\begingroup
\setlength{\parindent}{0pt}
\setlength{\parskip}{0.18em}
\renewcommand{\baselinestretch}{1.12}\selectfont
\hypersetup{colorlinks=true, linkcolor=blue, urlcolor=blue, citecolor=blue, pdfborder={0 0 0}}
\fontsize{12}{12}\selectfont
\newcommand{\apptoclinktext}[2]{\hyperref[#1]{\textcolor{blue}{#2}}}
\newcommand{\apptocentry}[2]{%
    \vspace{0.35em}%
    \noindent\hangindent=1.8em\hangafter=1%
    \apptoclinktext{#1}{#2}\hfill\apptoclinktext{#1}{\pageref{#1}}\par%
}
\newcommand{\apptocsubentry}[2]{%
    \vspace{0.35em}%
    \noindent\hspace*{1.5em}\hangindent=3.2em\hangafter=1%
    \apptoclinktext{#1}{#2}\hfill\apptoclinktext{#1}{\pageref{#1}}\par%
}

\begin{center}
{\Huge\bfseries\textsc{Appendix Contents}\par}
\vspace{1.2em}
\hrule height 0.6pt
\end{center}

\apptocentry{app:hyperparameters}{\textbf{A}\enspace Implementation Details and Hyperparameters}
\apptocsubentry{app:hyperparameters}{A.1\enspace High-Throughput Training Protocol}
\apptocsubentry{app:hyperparameters}{A.2\enspace Unified FastDSAC Hyperparameters}
\apptocsubentry{app:hyperparameters}{A.3\enspace Baseline Configurations}
\apptocsubentry{app:hyperparameters}{A.4\enspace Detailed Implementation Recipe}

\apptocentry{app:sim2real}{\textbf{B}\enspace Extended Details on Sim-to-Real Validation}

\apptocentry{app:fasttd3_dsact_background}{\textbf{C}\enspace Supplementary Background on FastTD3 and DSAC-T}

\apptocentry{app:algorithm}{\textbf{D}\enspace Architecture and Pseudocode of FastDSAC}

\apptocentry{app:related_work}{\textbf{E}\enspace Related Work}

\apptocentry{app:contemporary_comparison}{\textbf{F}\enspace Comparison with Contemporary Works}

\apptocentry{app:computational_resources}{\textbf{G}\enspace Computational Resources}

\apptocentry{app:wall_clock_time}{\textbf{H}\enspace Computational Efficiency and Wall-Clock Time}

\apptocentry{app:basketball_episode_heatmap}{\textbf{I}\enspace Heatmaps of Basketball Task}

\apptocentry{app:c51_support}{\textbf{J}\enspace C51 Support Sensitivity on \textit{Basketball}}

\apptocentry{sec:appendix_layernorm_ablation}{\textbf{K}\enspace Ablation Study on Layer Normalization}

\apptocentry{sec:appendix_target_entropy_ablation}{\textbf{L}\enspace Ablation Study on Target Entropy}

\apptocentry{app:full_results}{\textbf{M}\enspace Full Results}

\apptocentry{sec:limitations}{\textbf{N}\enspace Discussion of Limitations, Future Directions and Impact}

\vspace{0.5em}
\hrule height 0.6pt
\endgroup

\appendix
\onecolumn
\section{Implementation Details and Hyperparameters}
\label{app:hyperparameters}

\subsection{High-Throughput Training Protocol}
To ensure a rigorous and fair comparison, FastDSAC strictly inherits the high-throughput training protocol from FastTD3 \citep{seo2025fasttd3}. All domain-specific configurations, including parallel environment counts, replay buffer sizes, and update schedules (Update-to-Data (UTD) ratios), are matched to the official baseline settings. This protocol-level alignment ensures that any performance gains are attributable to the algorithmic innovations of FastDSAC rather than to discrepancies in computational budget or training schedules.

\begin{table}[H]
\centering
\caption{Domain-specific training protocols inherited from FastTD3~\citep{seo2025fasttd3}. These parameters are defined by the underlying high-throughput framework rather than being tuned for FastDSAC.}
\label{tab:protocol_settings}
\small
\begin{tabular}{lcccc}
\toprule
\textbf{Parameter} & \textbf{HumanoidBench} & \textbf{Isaac-Velocity} & \textbf{Isaac-General} & \textbf{MuJoCo PG} \\
\midrule
Parallel Environments ($N_{env}$) & 128 & 4,096 & 4,096 & 1,024 \\
Multi-step Return ($N_{steps}$) & 1 & 8 & 1 & 1 \\
Updates per Step ($N_{updates}$) & 2 & 4 & 2 & 2 \\
Discount Factor ($\gamma$) & 0.99 & 0.99 & 0.99 & 0.97 \\
Replay Buffer Size & $5 \times 10^4$ & $5 \times 10^3$ & $10^4$ & $10^4$ \\
AMP / AMP Dtype & True / bf16 & True / bf16 & True / bf16 & True / bf16 \\
Batch Size & 32,768 & 32,768 & 32,768 & 32,768 \\
\bottomrule
\end{tabular}
\end{table}

\subsection{Unified FastDSAC Hyperparameters}
In contrast to the per-task tuning often required by stochastic RL, FastDSAC utilizes a \textbf{unified algorithmic configuration} within each benchmark domain. As detailed in Table~\ref{tab:fastdsac_unified}, core parameters such as DEM temperature, target entropy, and log-std bounds are kept strictly constant across all tasks within a given domain. 

The differences in $\alpha_{init}$ and LayerNorm between HumanoidBench and lower-dimensional simulators (IsaacLab/MuJoCo PG) reflect basic adaptations to action dimensionality ($|\mathcal{A}|=61$ vs. $|\mathcal{A}| < 30$) and are consistent with standard literature on high-dimensional SAC \citep{ball2023efficient, nauman2024bigger}.

\begin{table}[H]
\centering
\caption{Unified FastDSAC algorithmic hyperparameters. These values define our core method and are fixed across all tasks within their respective domains.}
\label{tab:fastdsac_unified}
\begin{tabular}{l|c|c}
\toprule
\textbf{Hyperparameter} & \textbf{HumanoidBench} & \textbf{IsaacLab / MuJoCo PG} \\
\midrule
DEM Temperature ($\tau$) & 1.0 (Default) & 1.0 (Default) \\
Log-Std Max / Min & $1.0 \, / \, -10$ & $1.0 \, / \, -10$ \\
Target Entropy ($\mathcal{H}$) & 0 & 0 \\
Initial Alpha ($\alpha_{init}$) & 0.001 & 0.01 \\
Layer Normalization & Enabled & Disabled \\
Polyak Update Rate ($\tau_{polyak}$) & 0.005 & 0.005 \\
Weight Decay & $10^{-4}$ & $10^{-4}$ \\
AdamW Betas & $(0.9, 0.95)$ & $(0.9, 0.95)$ \\
Critic Stabilizer ($\epsilon$) & $10^{-6}$ & $10^{-6}$ \\
\bottomrule
\end{tabular}
\end{table}

\subsection{Baseline Configurations.} Results for the FastTD3 baseline are taken directly from the original paper \citep{seo2025fasttd3}. For the stochastic baseline, we use the FastSAC branch provided in the FastTD3 codebase and rerun it on the benchmark tasks considered in this paper. At the protocol level, FastSAC follows the same benchmark and task-specific setup as FastDSAC, including the FastTD3 replay/update schedule, number of training steps, replay buffer size, batch size, evaluation frequency, number of parallel environments, and task-specific discount factors. At the algorithm level, we largely retain the working configuration of this FastSAC implementation from the codebase, including the target-network soft update rate, weight decay, and log-standard-deviation bounds of $[-5, 2]$, while aligning a small number of generic optimizer settings with FastDSAC by using AdamW with $\beta=(0.9,0.95)$ and target entropy $\mathcal{H}=0$. LayerNorm is enabled on HumanoidBench and disabled on MuJoCo Playground and IsaacLab, matching the benchmark-specific usage in FastDSAC.

For the critic-ablation baseline \textit{FastSAC (C51$+$DEM)}, we retain the FastSAC training pipeline and replace the continuous critic with the same C51 critic family used in FastTD3, while adding DEM to the actor. All C51-specific hyperparameters follow FastTD3, including the support range and number of atoms. To keep the comparison controlled, we align shared hyperparameters such as weight decay, target entropy, and log-standard-deviation bounds ($[-10, 1]$) with FastDSAC, while leaving the remaining settings consistent with the inherited FastSAC/FastTD3 configurations. The DEM-specific settings follow the main FastDSAC setup, using a fixed temperature $\tau=1$ together with the same heterogeneity range $\beta_e \sim U[\beta_{\min}, \beta_{\max}]$. LayerNorm usage again follows the same benchmark-specific rule as above.

Overall, these baselines rigorously adhere to the stable configurations from their official codebases. We introduce only minimal, strictly controlled alignments (e.g., matching the target entropy) to ensure a fair comparison, explicitly avoiding any extensive per-baseline retuning that might introduce bias.

\subsection{Detailed Implementation Recipe}
\label{app:implementation_details}

To ensure stability and peak performance in the high-throughput continuous control regime, we adopt specific algorithmic and engineering choices grounded in recent literature and our empirical experiments.

\textbf{1. Layer Normalization (LayerNorm).} 
We apply LayerNorm in the actor and critic networks exclusively for the ultra-high-dimensional HumanoidBench domain ($|\mathcal{A}|=61$). Consistent with recent findings \citep{ball2023efficient,nauman2024bigger,seo2025learningsimtorealhumanoidlocomotion}, this regularization is critical for maintaining stable gradient flow in expansive action spaces. Conversely, for the MuJoCo Playground and IsaacLab tasks featuring smaller action spaces, FastDSAC achieves peak performance without LayerNorm. This contrast demonstrates that our performance gains stem primarily from core algorithmic mechanisms (DEM and continuous critic) rather than architectural dependencies. A comprehensive empirical analysis justifying these domain-specific choices is provided in Appendix~\ref{sec:appendix_layernorm_ablation}.

\textbf{2. Target Entropy.} 
Following recent work \citep{seo2025learningsimtorealhumanoidlocomotion}, we set the target entropy $\mathcal{H} = 0$ across all tasks. In high-dimensional spaces, the standard heuristic ($\mathcal{H} = -|\mathcal{A}|$) enforces aggressive variance decay, forcing the policy to prematurely drop its exploration budget and exacerbating the ``vanishing exploration'' problem. By setting $\mathcal{H} = 0$, we explicitly maintain a generous global exploration budget to prevent premature convergence (empirically validated in Appendix~\ref{sec:appendix_target_entropy_ablation}). While such high unmodulated variance would normally destabilize a complex humanoid, DEM structurally manages it. As analyzed in Section~\ref{sec:ablation:actor_ablation}, DEM safely offloads this excess entropy onto redundant joints, thereby preserving near-deterministic precision on critical actuators.

\textbf{3. Adaptive Temperature Optimization (auto-$\tau$).} 
The unified default DEM temperature ($\tau=1.0$) yields strong and robust performance across the vast majority of evaluated tasks and remains our primary recommendation for general use. However, a small subset of highly dynamic tasks requiring extreme whole-body coordination (e.g., \textit{Hurdle} and \textit{Balance Hard}) intrinsically benefit from uniform variance allocation (or exploration) across action dimension according to empirical experiments. To examine this potential without manual intervention, we introduce an adaptive variant, FastDSAC (auto-$\tau$), in which $\tau$ is treated as a network parameter and optimized end-to-end. Included in the main results (Figure~\ref{fig:main_results}) as an "upper bound" of DEM's autonomous exploration management. While the fixed-$\tau$ baseline remains the primary recommendation for general use because of its simplicity, this adaptive extension highlights the potential for future work to discover task-specific coordination patterns without relying on manual kinematic priors.

\textbf{4. High-Precision Critic Stability.} 
Unlike the original DSAC-T \citep{DuanDSACv2}, which requires a large numerical stabilizer (bias $= 0.1$) to prevent division-by-zero during continuous distribution modeling, the inherently stable statistics derived from our large-batch, high-throughput regime allow us to reduce this bias significantly to $10^{-6}$. This crucial reduction enables the FastDSAC critic to model states with low-variance return distributions with high fidelity, accurately capturing fine-grained reward structures that would otherwise be smoothed out by conservative numerical stabilization.

\textbf{5. Implementation Notes.}
In implementation, the temperature-scaled DEM logits are clipped to a bounded range before the Softmax for numerical stability, and the base log-standard deviation is mapped to fixed bounds via a tanh parameterization. For heterogeneous exploration, the environment-specific scaling factor is kept fixed within each episode and resampled only when the episode terminates, consistent with the mixed scaling exploration implementation inherited from FastTD3.

\begin{figure*}[!t]
     \centering
    \includegraphics[width=1\linewidth]{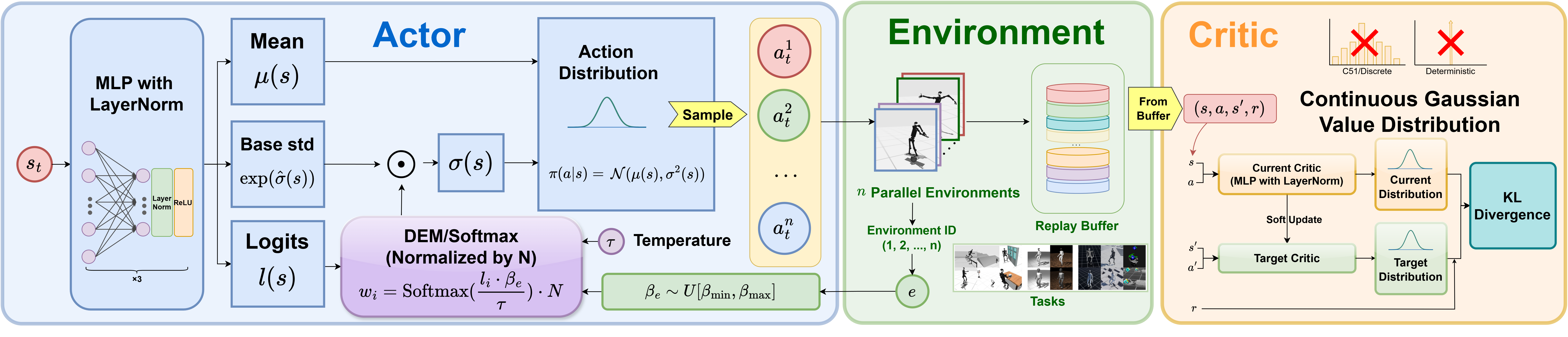}
    \caption{\textbf{Overview of the FastDSAC architecture.} \textbf{(Left) Actor with DEM:} The policy dynamically redistributes the exploration budget by modulating the base standard deviation $\hat{\sigma}_\phi(s)$ with weights $w_i$ (via element-wise multiplication $\odot$). These weights are derived from logits $l(s)$ and an environment-conditioned heterogeneity factor $\beta_e$ using a normalized Softmax. \textbf{(Middle) Environment:} Massively parallel environments collect experiences into a shared Replay Buffer for high-throughput training. \textbf{(Right) Continuous Critic:} The critic models the value function as a \textbf{continuous Gaussian distribution} to ensure accurate value estimation. It minimizes the \textbf{KL Divergence} between the current and target distributions, avoiding the quantization artifacts of discrete baselines (marked by $\times$).}
    \label{fig:architecture}
\end{figure*}

\section{Extended Details on Sim-to-Real Validation}
\label{app:sim2real}

To further examine whether the simulated performance of FastDSAC carries over to physical hardware, we deploy the learned policies on a Unitree G1 humanoid using the mature sim-to-real pipeline \citep{seo2025learningsimtorealhumanoidlocomotion}. This appendix summarizes the training pipeline, task setup, and deployment results behind the main-text results in Section~\ref{sec:sim2real}.

\textbf{Training Pipeline and Domain Randomization.} 
Policies are trained entirely in simulation using massive parallelization. To reduce the reality gap, the training pipeline applies domain randomization to key physical factors, including mass, friction, motor damping, latency, and external push perturbations. The final real-world performance should therefore be interpreted as the combined outcome of the strong FastDSAC method and the maturity of this underlying domain randomization pipeline.

\textbf{Task Formulations and Minimal Adaptation.} 
We evaluate two policy classes: (1) a joystick-conditioned locomotion policy (shown in the main text, Figure~\ref{fig:g1_sim2real}), and (2) a complex motion-tracking policy shown here. Crucially, the algorithmic adaptation is minimal. The core hyperparameters of FastDSAC (e.g., learning rates, DEM target entropy $\mathcal{H}=0$) remain identical to those used in the simulated IsaacLab (and MuJoCo Playground) experiments. We modify only environment-specific rollout settings to match the established deployment recipes, using $\gamma=0.99$ for motion tracking and $\gamma=0.97$ for locomotion, together with the corresponding multi-step return horizons and update-to-data ratios.

\begin{figure}[h]
    \centering
    \begin{subfigure}[b]{0.48\textwidth}
        \centering
        \includegraphics[width=\linewidth]{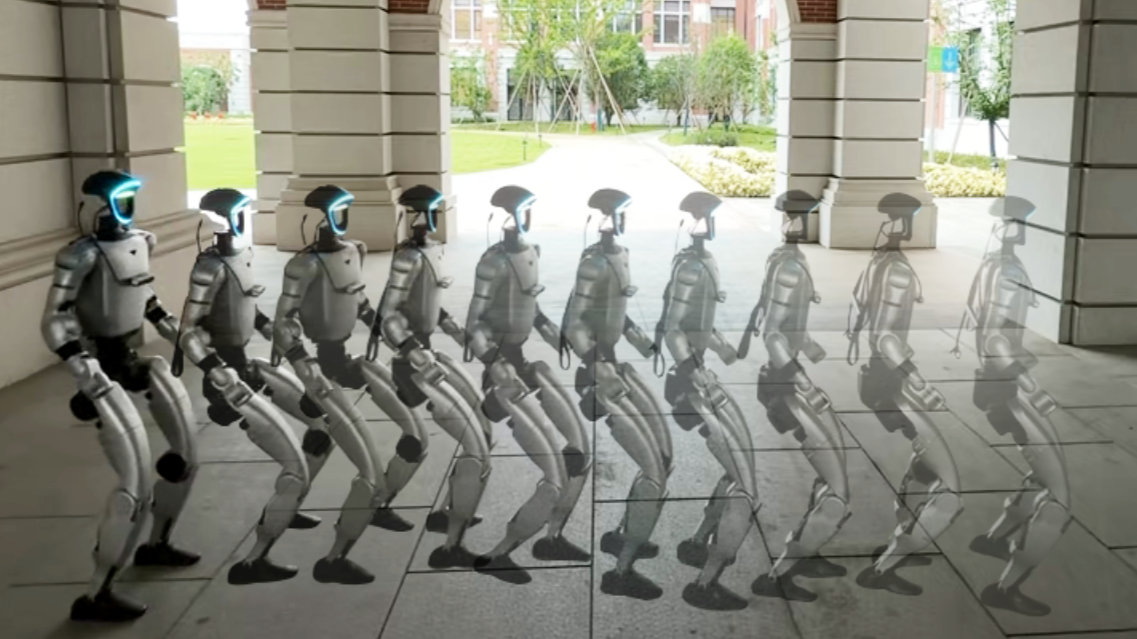}
        \caption{Joystick-conditioned backward walking}
    \end{subfigure}
    \hfill
    \begin{subfigure}[b]{0.48\textwidth}
        \centering
        \includegraphics[width=\linewidth]{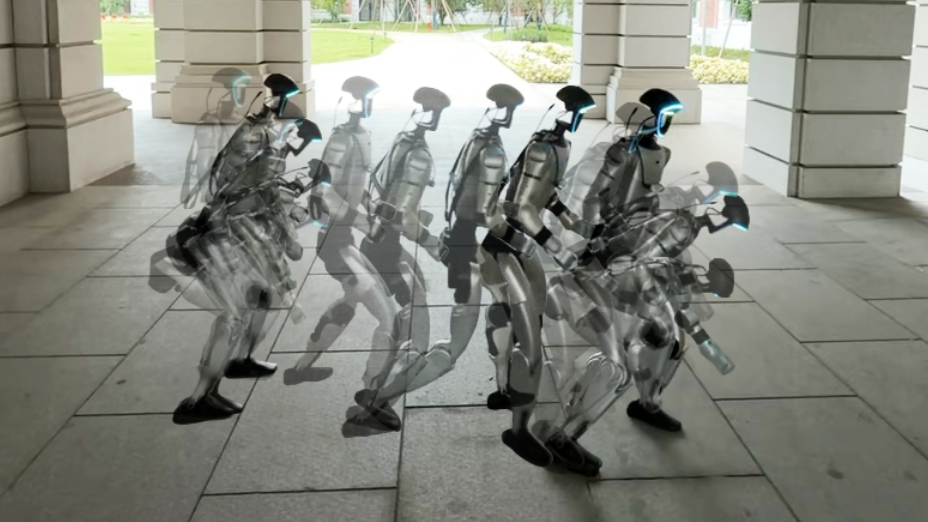}
        \caption{``Squat-carry-walk-squat'' motion tracking}
    \end{subfigure}
    \caption{\textbf{Additional Sim-to-Real Results.} Time-lapse sequences of zero-shot FastDSAC policies deployed on the Unitree G1. The robot executes backward walking under real-time joystick commands and a ``squat-carry-walk-squat'' motion-tracking sequence.}
    \label{fig:g1_locomotion}
\end{figure}

\textbf{Deployment Results.}
The learned policies run directly on the G1's onboard computer. For the locomotion task, the robot tracks real-time joystick commands; for motion tracking, it follows the reference ``squat-carry-walk-squat'' sequence. Figure~\ref{fig:g1_locomotion} presents time-lapse sequences of backward walking and the motion-tracking policy. Both policies show strong zero-shot stability and remain stable under moderate physical push perturbations. These observations are consistent with the main-text result that policies trained with FastDSAC in simulation can transfer to robust real-world control.

\section{Supplementary Background on FastTD3 and DSAC-T}
\label{app:fasttd3_dsact_background}

\subsection{FastTD3 Recipe Inherited by FastDSAC}
FastTD3 is presented as a simplified high-throughput TD3 recipe built from parallel simulation, large-batch updates, a distributional critic, and tuned hyperparameters~\citep{seo2025fasttd3}. The original paper emphasizes that massively parallel environments primarily improve wall-clock efficiency and also increase data diversity, which helps deterministic policy gradients compensate for weak single-trajectory exploration. It further reports that an unusually large batch size of $32{,}768$ stabilizes critic learning under massively parallel rollouts, and that the resulting diversity can reduce the effective off-policyness of updates enough to keep training stable even without residual connections or LayerNorm.

For value estimation, FastTD3 adopts a discrete distributional critic from the C51 family:
\begin{equation}
    Z^{\mathrm{C51}}_{\theta}(s,a)=\sum_{i=1}^{N} p_{\theta,i}(s,a)\,\delta_{z_i},
\end{equation}
where $\{z_i\}_{i=1}^{N}$ is a fixed support spanning $[v_{\min},v_{\max}]$, $p_{\theta,i}(s,a)$ is the predicted probability of atom $z_i$, and $\delta_{z_i}$ denotes a Dirac mass. Given the target action from the TD3 target actor, the distributional Bellman target is
\begin{equation}
    \mathcal{T}Z^{\mathrm{C51}}_{\theta'}(s,a)=r(s,a)+\gamma Z^{\mathrm{C51}}_{\theta'}(s',a'),
\end{equation}
which is projected back onto the fixed support to obtain $\boldsymbol{m}=\Phi\mathcal{T}Z^{\mathrm{C51}}_{\theta'}$. The critic is then trained by the Cross-Entropy objective
\begin{equation}
    \mathcal{L}_{\mathrm{C51}}(\theta)=-\sum_{i=1}^{N} m_i \log p_{\theta,i}(s,a).
\end{equation}
This discrete critic is helpful in the original FastTD3 study, but it introduces support-related hyperparameters $(v_{\min},v_{\max})$ and keeps value learning tied to a fixed atomization of returns. FastDSAC retains the same high-throughput protocol and replay/update structure, but replaces this critic family with a continuous Gaussian critic to avoid support tuning and quantization artifacts.

Beyond the critic parameterization, the FastTD3 recipe includes two implementation patterns that are inherited conceptually in FastDSAC. First, FastTD3 studies a mixed exploration-noise schedule across environments,
\begin{equation}
    a_t^{(e)}=\mu_{\phi}(s_t^{(e)})+\epsilon_t^{(e)}, \qquad \epsilon_t^{(e)}\sim\mathcal{N}(0,\sigma_e^2 I), \qquad \sigma_e\sim U[\sigma_{\min},\sigma_{\max}],
\end{equation}
and reports that although the mixed-noise scheme itself does not produce a major gain, it remains a cheap way to expose parallel environments to diverse exploration scales. FastDSAC inherits this spirit through heterogeneous DEM scaling factors $\beta_e$, while replacing global noise-scale variation with structured, dimension-wise variance redistribution. Second, FastTD3 keeps the optimization recipe deliberately simple: standard MLP backbones, low update-to-data ratios, and benchmark-specific parallel rollout protocols summarized in Appendix~\ref{app:hyperparameters}. FastDSAC preserves this training backbone so that the main algorithmic difference remains the actor-side exploration structure and the critic parameterization.

\subsection{DSAC-T Components Used in FastDSAC}
\label{app:dsact_components}
For completeness, we summarize only the DSAC-T ingredients directly relevant to FastDSAC's critic design. DSAC and DSAC-T model the return distribution as a continuous Gaussian, so the critic directly predicts both the mean Q-value and its uncertainty scale~\citep{Duan2020DistributionalSA,DuanDSACv2}:
\begin{equation}
    Z_{\theta}(\cdot|s,a)=\mathcal{N}(Q_{\theta}(s,a),\sigma_{\theta}^{2}(s,a)).
\end{equation}
Given $a'\sim\pi_{\phi}(\cdot|s')$ and a target sample $z'\sim Z_{\theta'}(s',a')$, the standard distributional target is
\begin{equation}
    y_z=r+\gamma\bigl(z'-\alpha\log\pi_{\phi}(a'|s')\bigr).
\end{equation}
DSAC begins from a sample-based negative log-likelihood objective for this target return under the current Gaussian critic,
\begin{equation}
    J_Z(\theta)=-\mathbb{E}\Big[\log \mathcal{P}\bigl(y_z \mid Z_{\theta}(\cdot|s,a)\bigr)\Big].
\end{equation}
Under the Gaussian parameterization above, this objective is equivalent, up to constants independent of $\theta$, to
\begin{equation}
    J_Z(\theta)=\mathbb{E}\left[
    \frac{\bigl(y_z-Q_{\theta}(s,a)\bigr)^2}{2\sigma_{\theta}^{2}(s,a)}
    +\log \sigma_{\theta}(s,a)
    \right].
\end{equation}
DSAC decomposes the resulting critic gradient into mean-related and variance-related parts, and stabilizes the latter by clipping the random target return:
\begin{equation}
    C(y_z;b)={\rm clip}\left(y_z,Q_{\theta}(s,a)-b,Q_{\theta}(s,a)+b\right).
\end{equation}
This yields the clipped gradient form
\begin{equation}
\label{eq:app_dsacv1_grad}
\begin{aligned}
\nabla_{\theta}J_{Z}\approx
\mathbb{E}\Big[
&-\frac{y_z-Q_{\theta}(s,a)}{\sigma_{\theta}(s,a)^2}\nabla_{\theta}Q_{\theta}(s,a) \\
&-\frac{\big(C(y_z;b)-Q_{\theta}(s,a)\big)^2-\sigma_{\theta}(s,a)^2}{\sigma_{\theta}(s,a)^3}\nabla_{\theta}\sigma_{\theta}(s,a)
\Big].
\end{aligned}
\end{equation}

\paragraph{Expected Value Substituting.}
DSAC-T observes that the first term in Eq.~\eqref{eq:app_dsacv1_grad} still inherits the randomness of $y_z$. It therefore introduces the target Q-value
\begin{equation}
    y_q=r+\gamma\bigl(Q_{\theta'}(s',a')-\alpha\log\pi_{\phi}(a'|s')\bigr),
\end{equation}
which satisfies $\mathbb{E}[y_z\mid a'] = y_q$. Replacing the first occurrence of $y_z$ by $y_q$ gives
\begin{equation}
\label{eq:app_dsact_evs}
\begin{aligned}
\nabla_{\theta}J_{Z}\approx
\mathbb{E}\Big[
&-\frac{y_q-Q_{\theta}(s,a)}{\sigma_{\theta}(s,a)^2}\nabla_{\theta}Q_{\theta}(s,a) \\
&-\frac{\big(C(y_z;b)-Q_{\theta}(s,a)\big)^2-\sigma_{\theta}(s,a)^2}{\sigma_{\theta}(s,a)^3}\nabla_{\theta}\sigma_{\theta}(s,a)
\Big].
\end{aligned}
\end{equation}
This refinement makes the mean update closer to the non-distributional SAC target and reduces gradient randomness in the critic update.

\paragraph{Twin Value Distribution Learning.}
The second refinement is a distributional analogue of clipped double Q-learning. DSAC-T trains two Gaussian critics $Z_{\theta_1}$ and $Z_{\theta_2}$ independently and selects the target critic with the smaller mean:
\begin{equation}
    j^{*}:=\arg\min_{j\in\{1,2\}} Q_{\theta'_j}(s',a').
\end{equation}
The target return and target Q-value are then defined as
\begin{equation}
\label{eq:app_dsact_twin_targets}
\begin{aligned}
    y_{z}^{\min} &= r+\gamma\bigl(z'_{j^{*}}-\alpha\log\pi_{\phi}(a'|s')\bigr), \qquad z'_{j^{*}}\sim Z_{\theta'_{j^{*}}}(s',a'), \\
    y_{q}^{\min} &= r+\gamma\bigl(Q_{\theta'_{j^{*}}}(s',a')-\alpha\log\pi_{\phi}(a'|s')\bigr).
\end{aligned}
\end{equation}
Substituting Eq.~\eqref{eq:app_dsact_twin_targets} into Eq.~\eqref{eq:app_dsact_evs} yields the twin-distribution critic update, and the actor objective becomes
\begin{equation}
    J_{\pi}(\phi)=\mathbb{E}_{s\sim\mathcal{B},a\sim\pi_{\phi}}\bigl[\min_{j\in\{1,2\}}Q_{\theta_j}(s,a)-\alpha\log\pi_{\phi}(a|s)\bigr].
\end{equation}
This refinement reduces overestimation bias by building both critic and actor targets from the more conservative value distribution.

\paragraph{Variance-Based Critic Gradient Adjustment.}
The third refinement removes the fixed clipping boundary and the reward-scale sensitivity of the mean update. DSAC-T sets the clipping boundary adaptively as
\begin{equation}
    b=\xi \mathbb{E}_{(s,a)\sim\mathcal{B}}\big[\sigma_{\theta}(s,a)\big],
\end{equation}
where $\xi$ is a constant controlling the clipping range (typically set $\xi =3$, following the three-sigma rule), and introduces the variance-based scaling weight
\begin{equation}
    \omega=\mathbb{E}_{(s,a)\sim\mathcal{B}}\big[\sigma_{\theta}(s,a)^2\big].
\end{equation}
Using moving averages $b_j$ and $\omega_j$ for the two critics, the DSAC-T gradient becomes
\begin{equation}
\label{eq:app_dsact_final}
\begin{aligned}
\nabla_{\theta_j}J_{Z}^{\mathrm{scale}}\approx
(\omega_j+\epsilon_{\omega})\mathbb{E}\Big[
&-\frac{y_{q}^{\min}-Q_{\theta_j}(s,a)}{\sigma_{\theta_j}(s,a)^2+\epsilon}\nabla_{\theta_j}Q_{\theta_j}(s,a) \\
&-\frac{\big(C(y_{z}^{\min};b_j)-Q_{\theta_j}(s,a)\big)^2-\sigma_{\theta_j}(s,a)^2}{\sigma_{\theta_j}(s,a)^3+\epsilon}\nabla_{\theta_j}\sigma_{\theta_j}(s,a)
\Big].
\end{aligned}
\end{equation}
The corresponding moving-average updates are
\begin{equation}
\begin{aligned}
    b_j &\leftarrow \tau\,\xi\,\mathbb{E}_{(s,a)\sim\mathcal{B}}\big[\sigma_{\theta_j}(s,a)\big]+(1-\tau)b_j, \\
    \omega_j &\leftarrow \tau\,\mathbb{E}_{(s,a)\sim\mathcal{B}}\big[\sigma_{\theta_j}(s,a)^2\big]+(1-\tau)\omega_j.
\end{aligned}
\end{equation}

FastDSAC keeps the first two refinements because they are directly compatible with our high-throughput setting. For the third, we only retain the gradient scaling idea. In the final implementation, Section~\ref{sec:method_critic} removes the dynamic clipping boundary, uses an unclipped sampled-target variance update, and reduces the numerical stabilizer to better resolve low-variance states in the large-batch regime.

\section{Architecture and Pseudocode of FastDSAC}
We provide the complete pseudocode for FastDSAC in Algorithm 1. For FastDSAC architecture, please refer to Figure~\ref{fig:architecture}.
\label{app:algorithm}
\begin{algorithm}[H]
    \caption{FastDSAC}
    \label{alg:fastdsac}
    \begin{algorithmic}[1]
        \REQUIRE $n$ parallel environments, replay buffer $\mathcal{B}$, updates per step $U$, discount $\gamma$, Polyak coefficient $\tau_{polyak}$, DEM temperature $\tau$, heterogeneity range $[\beta_{\min},\beta_{\max}]$, target entropy $\mathcal{H}$.
        \ENSURE Learned policy parameters $\phi$.
        \STATE Initialize actor $\pi_\phi$ with output heads $(\mu_\phi,\hat{\sigma}_\phi,l_\phi)$, twin critics $Z_{\theta_j}(s,a) \sim \mathcal{N}(Q_{\theta_j}(s,a),\sigma_{\theta_j}^2(s,a))$, target parameters $\theta'_j \gets \theta_j$, entropy coefficient $\alpha$, and $\mathcal{B} \gets \emptyset$.
        \FOR{\text{each training iteration}}
            \STATE \textbf{Parallel Rollout:}
            \FOR{environment $e \in [n]$ \textbf{in parallel}}
                \STATE Sample heterogeneity factor $\beta_e \sim \mathcal{U}[\beta_{\min},\beta_{\max}]$, observe $s_t^{(e)}$.
                \STATE Forward pass: $(\mu, \hat{\sigma}, l) \gets \pi_\phi(s_t^{(e)})$.
                \STATE Compute dimension-wise weights: $w_e \gets N \cdot \mathrm{Softmax}(\beta_e l / \tau)$ per Eq.~\eqref{eq:weights}.
                \STATE Modulate exploration: $\sigma_e \gets w_e \odot \exp(\hat{\sigma})$.
                \STATE Sample noise $\xi \sim \mathcal{N}(0,I)$, execute action $a_t^{(e)} \gets \mu + \sigma_e \odot \xi$.
                \STATE Collect transition $(s_t^{(e)}, a_t^{(e)}, r_t^{(e)}, s_{t+1}^{(e)}, d_t^{(e)})$, store in $\mathcal{B}$.
            \ENDFOR
            \STATE \textbf{Distributional Soft Policy Evaluation \& Improvement:}
            \FOR{gradient step $u \in [U]$}
                \STATE Sample batch $(s,a,r,s',d) \sim \mathcal{B}$, compute $\bar{\gamma} \gets \gamma(1-d)$.
                \STATE \textit{Evaluation:} Sample next action $a' \sim \pi_\phi(\cdot|s')$, compute target $y_q^{\min} \gets r + \bar{\gamma}(\min_{j \in \{1,2\}} Q_{\theta'_j}(s',a') - \alpha \log \pi_\phi(a'|s'))$.
                \FOR{$j \in \{1,2\}$}
                    \STATE Sample $z'_j \sim Z_{\theta'_j}(s',a')$, compute $y_{z,j} \gets r + \bar{\gamma} \bigl(z'_{j}-\alpha\log\pi_{\phi}(a'|s')\bigr)$.
                    \STATE Compute critic scaling: $\omega_j \gets \mathbb{E}_{\mathcal{B}}[\sigma_{\theta_j}^2(s,a)]$.
                    \STATE Update critic: $\theta_j \gets \theta_j - \eta_Q \nabla_{\theta_j} J_Z$ via Eq.~\eqref{eq:fastdsac_critic_grad} with $(y_q^{\min}, y_{z,j}, \omega_j)$.
                \ENDFOR
                \STATE \textit{Improvement:} Sample action $a \sim \pi_\phi(\cdot|s)$ from batch.
                \STATE Update actor: $\phi \gets \phi - \eta_\pi \nabla_\phi \mathcal{L}_{\text{actor}}$ where $\mathcal{L}_{\text{actor}} = \mathbb{E}[\alpha \log \pi_\phi(a|s) - Q_\theta(s,a)]$.
                \STATE Optimize entropy coefficient: $\alpha \gets \alpha - \eta_\alpha \mathbb{E}[-\log \pi_\phi(a|s) - \mathcal{H}]$.
                \STATE Target update: $\theta'_j \gets (1-\tau_{polyak}) \theta'_j + \tau_{polyak} \theta_j$ for $j \in \{1,2\}$.
            \ENDFOR
        \ENDFOR
    \end{algorithmic}
\end{algorithm}

\section{Related Work}\label{app:related_work}
\textbf{Scalable Reinforcement Learning for High-Dimensional Control.} Recent advances in physics simulation have shifted RL for high-dimensional humanoid control toward massively parallel training regimes \citep{rudin2022learning,mittal2025isaaclab,zakka2025mujoco}. In this high-throughput setting, deterministic off-policy methods such as Parallel Q-Learning (PQL) \citep{li2023parallel} and FastTD3 \citep{seo2025fasttd3} have been especially successful, largely because their training dynamics remain stable under aggressive replay-based updates in large action spaces. More recently, stochastic off-policy RL has also begun to benefit from this regime. For example, FlashSAC \citep{kim2026flashsac} demonstrates that SAC can be scaled effectively through recipe-level stabilization and improved system throughput. However, its evidence on HumanoidBench is primarily concentrated on lower-dimensional H1 locomotion tasks, whereas our focus is the considerably more challenging 61-dimensional H1Hand whole-body control setting, evaluated across a comprehensive suite of 25 diverse tasks.

Model-based approaches such as DreamerV3 \citep{hafner2023mastering}, TD-MPC2 \citep{hansen2023td}, and TDMPBC \citep{zhuang2025tdmpbc} can achieve strong sample efficiency, but they typically introduce substantial computational overhead from world-model learning and inference, often leading to longer wall-clock training in large-scale continuous-control settings. By contrast, model-free off-policy methods remain attractive in massively parallel environments because they better exploit simulator throughput.

From a methodological perspective, Maximum Entropy RL methods such as SAC offer appealing exploration and robustness properties \citep{haarnoja2018soft}, but their effectiveness deteriorates sharply as action dimensionality grows. In practice, applying SAC to domains with $|\mathcal{A}|>60$ has remained challenging due to the curse of dimensionality, often requiring substantial architectural modification and still frequently underperforming strong deterministic baselines on the most difficult tasks \citep{obandoceron2025simplicialembeddingsimprovesample,seo2025fasttd3,kim2026flashsac}. FastDSAC is designed specifically for this regime. Rather than relying solely on scaling heuristics or abandoning stochasticity in favor of deterministic actors, FastDSAC directly targets the two core bottlenecks that hinder stochastic off-policy RL in very high-dimensional control: ineffective exploration and unstable value estimation. FastDSAC fills this gap by making stochastic maximum entropy RL practical in the high-throughput, $61$-DoF control regime, where prior success has been dominated by deterministic methods.

\textbf{Maximum Entropy RL and Value Estimation in High Dimensions.} Scaling Soft Actor-Critic (SAC) to high-dimensional action spaces introduces two fundamental challenges: inefficient exploration and value overestimation.

First, standard Gaussian policies suffer from \emph{vanishing exploration} \citep{wei2026scalableexplorationhighdimensionalcontinuous}, where the effective exploration volume collapses as dimensionality increases. Moreover, they allocate variance uniformly across both task-relevant and redundant action dimensions, resulting in poorly directed exploration \citep{zhang2024on,sferrazza2024humanoidbench}. Recent works address this issue from different perspectives. QFLEX \citep{wei2026scalableexplorationhighdimensionalcontinuous} mitigates the issue through flow-based action modeling. ERA \citep{kang2025entropyregularizingactivationboosting} proposes a different route by enforcing a minimum sampling-entropy threshold through output-layer activations, and reports gains across multiple continuous-control algorithms, including FastSAC-ERA on HumanoidBench. FastDSAC is related to ERA in spirit, since both methods reshape how stochasticity is expressed by the actor, but the underlying mechanisms differ. ERA imposes an explicit minimum-entropy constraint through output-layer activations, whereas our Dimension-wise Entropy Modulation (DEM) autonomously redistributes the exploration variance across specific dimensions based on task demands, avoiding the need for rigid, hard-coded entropy thresholds.

Second, accurate Q-function estimation becomes increasingly difficult in high-dimensional spaces due to severe extrapolation errors on out-of-distribution (OOD) actions \citep{Bhatt2019CrossQBN,neumann2024vlearnoffpolicylearning}. This often results in \emph{Q-bias explosion} \citep{Bhatt2019CrossQBN} and pronounced overestimation, as the policy progressively over-exploits artifacts in the critic approximation \citep{seo2025coarsetofineqnetworkactionsequence}. To mitigate this issue, prior methods such as TQC \citep{kuznetsov2020controllingoverestimationbiastruncated} and FastTD3 \citep{seo2025fasttd3} adopt distributional RL with truncated quantiles or categorical atoms (C51), respectively. While effective, these discrete approximations depend on manually specified value supports and a finite number of atoms, which can be brittle under varying reward scales and may introduce quantization error that limits value precision.

In contrast, FastDSAC employs a Continuous Gaussian Critic \citep{DuanDSACv2}. Rather than truncating the upper tail as in TQC, our critic models the full return distribution in a continuous form, thereby avoiding fixed-support discretization. Furthermore, by disentangling the gradient estimators, the continuous critic naturally suppresses mean updates on highly uncertain actions through inverse-variance scaling, providing a principled mechanism for reducing overestimation in high-dimensional settings.

\section{Comparison with Contemporary Works}
\label{app:contemporary_comparison}

\begin{figure}[h]
    \centering
    \begin{subfigure}{0.48\textwidth}
        \centering
        \includegraphics[width=\linewidth]{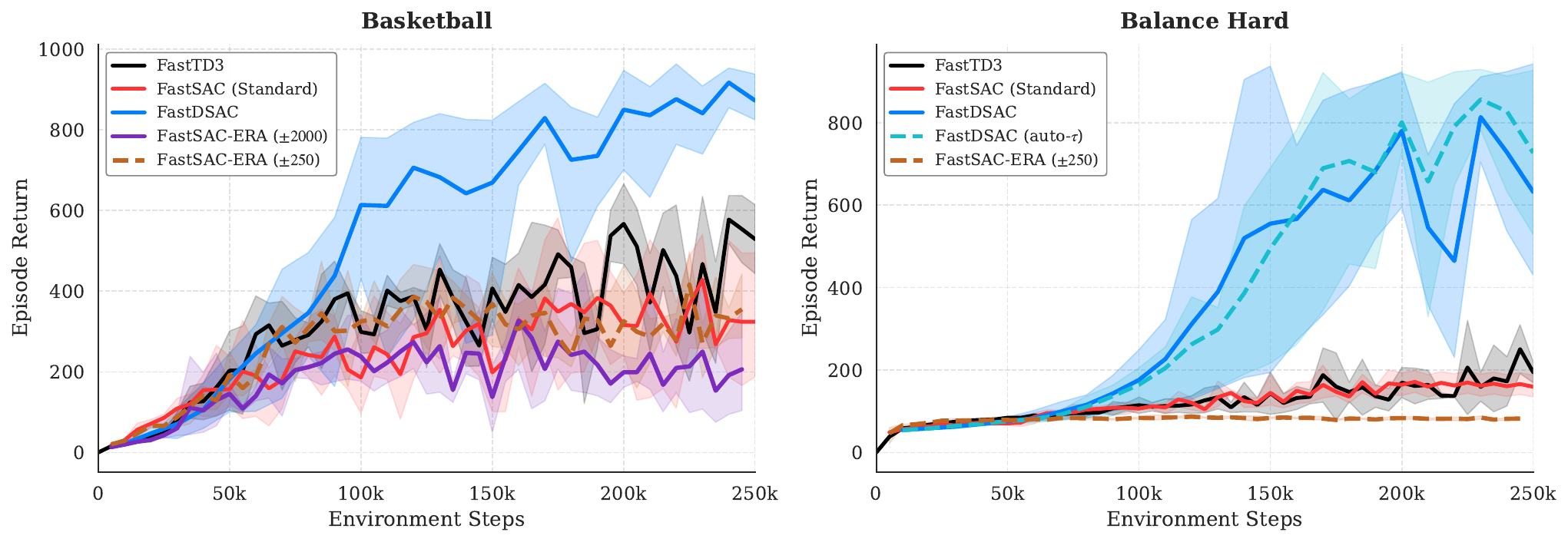}
        \caption{Comparison with FastSAC-ERA}
        \label{fig:era_comp}
    \end{subfigure}
    \hfill
    \begin{subfigure}{0.48\textwidth}
        \centering
        \includegraphics[width=\linewidth]{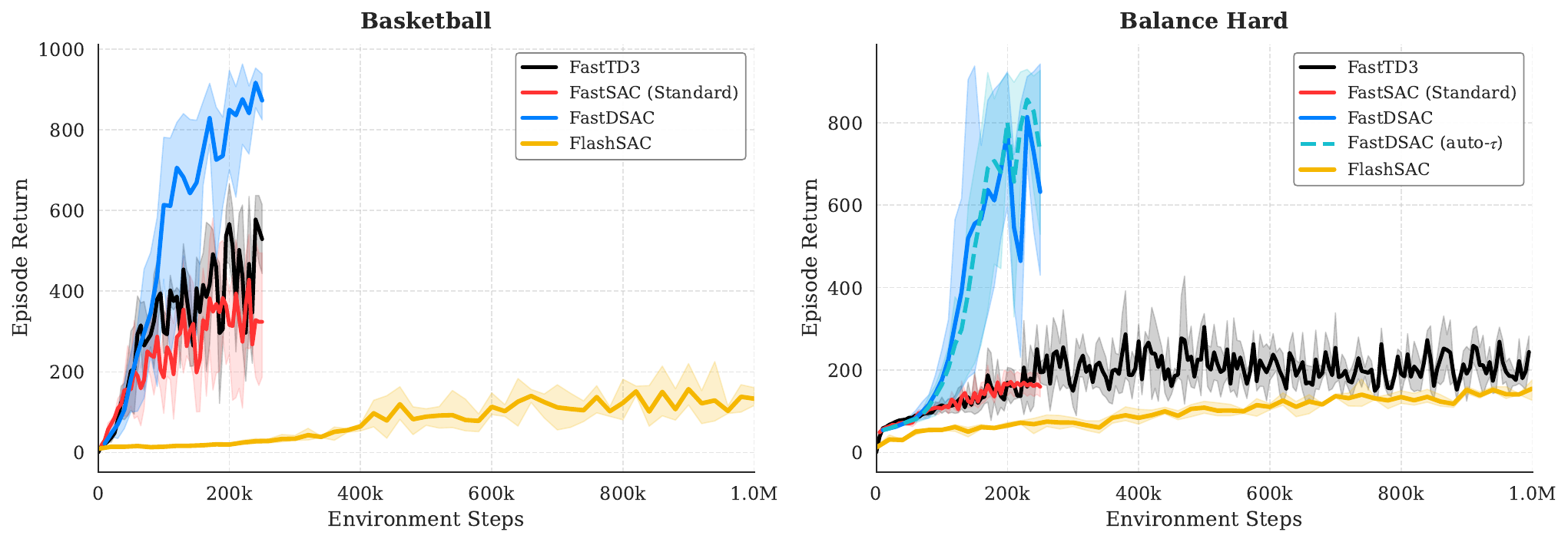}
        \caption{Comparison with FlashSAC}
        \label{fig:flashsac_comp}
    \end{subfigure}
    \caption{Performance comparison against concurrent baselines on 61-DoF HumanoidBench tasks. \textbf{(a)} FastSAC-ERA plateaus early, as global entropy thresholds restrict localized variance pruning. \textbf{(b)} FlashSAC struggles to scale to the 61-DoF regime despite extended training budgets, highlighting the necessity of algorithmic exploration mechanisms.}
    \label{fig:contemporary_comparison}
\end{figure}

To strictly assess the scalability of FastDSAC, we compare it against two recent concurrent methods, FastSAC-ERA \citep{kang2025entropyregularizingactivationboosting} and FlashSAC \citep{kim2026flashsac}. We evaluate these baselines on the challenging 61-DoF \textit{Basketball} and \textit{Balance Hard} tasks. The learning curves are presented in Figure~\ref{fig:contemporary_comparison}.

\textbf{Comparison with FastSAC-ERA.} 
ERA maintains exploration by enforcing a global minimum entropy threshold. However, as shown in Figure~\ref{fig:contemporary_comparison}(a), FastSAC-ERA plateaus early at suboptimal returns (e.g., $\approx 100$ in \textit{Balance Hard}). This underperformance highlights a fundamental limitation of global entropy constraints in very high-dimensional control: enforcing a uniform entropy lower bound prevents the localized variance pruning required for precision-critical joints. Successfully coordinating a 61-DoF system requires selectively offloading variance to redundant joints, a structure that rigid global thresholds inherently disrupt.

\textbf{Comparison with FlashSAC.} 
FlashSAC improves wall-clock efficiency through system-level throughput optimizations and recipe-level heuristics, demonstrating strong results on lower-dimensional locomotion ($|\mathcal{A}| \approx 20$). However, when applied to the 61-DoF regime shown in Figure~\ref{fig:contemporary_comparison}(b), it struggles to learn effective behaviors, scoring below 200 even after $10^6$ environment steps. While FlashSAC's system engineering is valuable for simulation performance and real robot deployment, its performance drop in higher dimensions suggests that raw throughput cannot substitute for explicit algorithmic mechanisms (such as DEM) needed to overcome the curse of dimensionality.

\section{Computational Resources}
\label{app:computational_resources}
We conducted our experiments on a heterogeneous computing cluster consisting of high-performance workstations. The experiments were distributed across two primary hardware configurations:

(1) \textbf{Configuration A:} Equipped with an Intel Core i7-14700K CPU (20 physical cores, 28 logical threads) and 128 GB of system RAM. Accelerated computing was provided by a single NVIDIA GeForce RTX 4090. 

(2) \textbf{Configuration B:} Equipped with an AMD Ryzen 9 9950X CPU (16 physical cores, 32 logical threads) and 192 GB of system RAM. This node features an NVIDIA RTX 6000 (Blackwell Workstation Edition) with 96 GB of VRAM, facilitating large-scale parallel training. 

\textbf{Resource Allocation.} To ensure efficient training and accommodate varying memory requirements, the memory-intensive IsaacLab environments, specifically \textit{Isaac-Velocity-Flat-G1-v0}, \textit{Isaac-Velocity-Rough-G1-v0}, \textit{Isaac-Repose-Cube-Allegro-Direct-v0}, and \textit{Isaac-Repose-Cube-Shadow-Direct-v0}, were trained on \textbf{Configuration B} to leverage its extensive VRAM capacity. All other tasks, including the HumanoidBench and MuJoCo Playground suites, were trained on \textbf{Configuration A}. This hardware setup ensured efficient training on high-dimensional tasks, with each experiment converging in under 5 hours.

\section{Computational Efficiency and Wall-Clock Time}
\label{app:wall_clock_time}

\begin{figure}[ht]
    \centering
    \includegraphics[width=\linewidth]{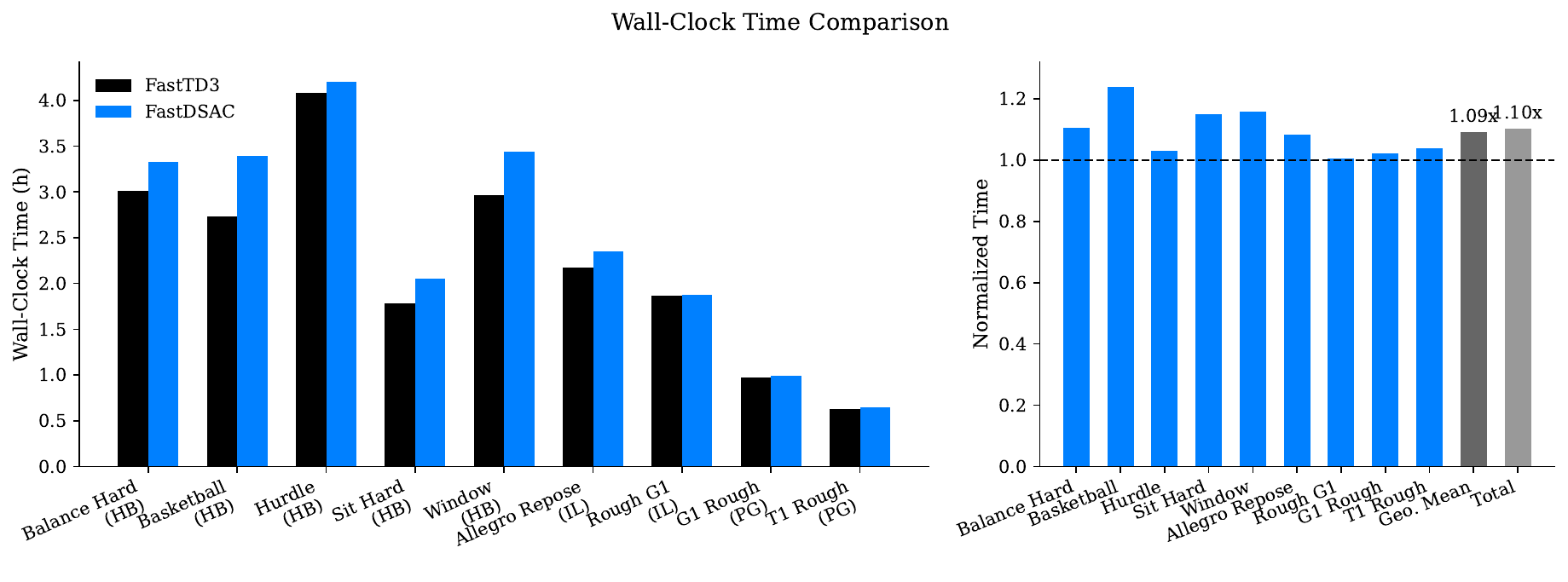}
    \caption{\textbf{Wall-clock time comparison between FastDSAC and FastTD3.} Evaluated on representative tasks from HumanoidBench, IsaacLab, and MuJoCo Playground. Left: absolute per-task total training time (hours). Right: relative wall-clock time normalized against FastTD3, detailing per-task ratios, the geometric-mean ratio, and the overall total-time ratio across all nine tasks.}
    \label{fig:wallclock_probe_extended_comparison}
\end{figure}

We evaluate the wall-clock training time of FastDSAC against FastTD3 across nine representative tasks from HumanoidBench, IsaacLab, and MuJoCo Playground (Figure~\ref{fig:wallclock_probe_extended_comparison} and Table~\ref{tab:wallclock_probe_extended}). To ensure a strictly controlled evaluation, both algorithms were benchmarked locally on identical hardware (a single NVIDIA GeForce RTX 4090 GPU and an Intel Core i7-14700K CPU). This hardware configuration naturally yields different absolute timings compared to the A100 GPUs used in the original FastTD3 publication.

Across all nine tasks, FastTD3 requires 20.21 hours in total, whereas FastDSAC requires 22.28 hours. This corresponds to an overall relative time ratio of 1.10$\times$ (geometric mean: 1.09$\times$). These results demonstrate that FastDSAC maintains comparable computational efficiency to the highly optimized FastTD3 baseline. Crucially, FastDSAC delivers significantly stronger performance and stability with only a marginal $\approx 10\%$ overhead in training time.

\begin{table}[ht]
\centering
\small
\setlength{\tabcolsep}{4pt}
\begin{tabular}{llcccc}
\toprule
Benchmark & Task & Steps & FastTD3 (h) & FastDSAC (h) & Rel. Time \\
\midrule
HB & Balance Hard & 250K & 3.01 & 3.33 & 1.10$\times$ \\
HB & Basketball & 250K & 2.73 & 3.39 & 1.24$\times$ \\
HB & Hurdle & 300K & 4.08 & 4.21 & 1.03$\times$ \\
HB & Sit Hard & 150K & 1.78 & 2.05 & 1.15$\times$ \\
HB & Window & 250K & 2.96 & 3.44 & 1.16$\times$ \\
IL & Allegro Repose & 200K & 2.17 & 2.35 & 1.08$\times$ \\
IL & Rough G1 & 100K & 1.87 & 1.88 & 1.01$\times$ \\
PG & G1 Rough & 45K & 0.97 & 0.99 & 1.02$\times$ \\
PG & T1 Rough & 45K & 0.63 & 0.65 & 1.03$\times$ \\
\midrule
HB & Geo. mean & -- & -- & -- & 1.13$\times$ \\
IL & Geo. mean & -- & -- & -- & 1.04$\times$ \\
PG & Geo. mean & -- & -- & -- & 1.03$\times$ \\
All & Geo. mean & -- & -- & -- & 1.09$\times$ \\
All & Total & -- & 20.21 & 22.28 & 1.10$\times$ \\
\bottomrule
\end{tabular}
\caption{\textbf{Wall-clock time comparison on representative HumanoidBench, IsaacLab, and MuJoCo Playground tasks.} All timings were measured locally on a single RTX 4090 GPU.}
\label{tab:wallclock_probe_extended}
\end{table}

\newpage
\section{Heatmaps of \textit{Basketball} Task}
\label{app:basketball_episode_heatmap}

\begin{figure}[H]
    \centering
    \includegraphics[width=0.85\linewidth]{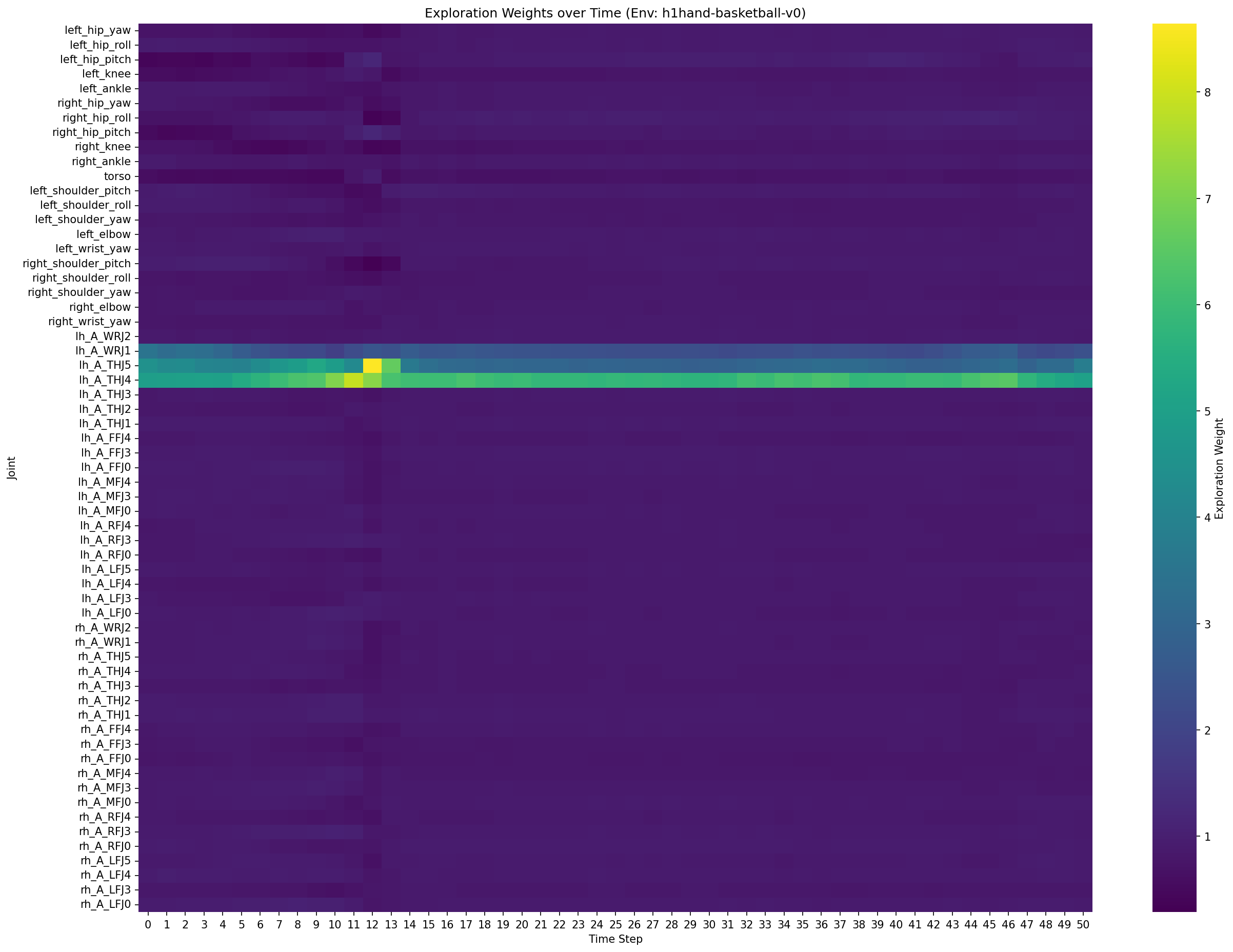}
    \caption{\textbf{Exploration weights ($w_i$) during a Basketball episode.} High weights (bright) concentrate on the left thumb and wrist, while lower weights (dark) on the legs, torso, and right arm.}
    \label{fig:heatmap_basketball}
\end{figure}

\begin{figure}[H]
	\centering
	\includegraphics[width=0.85\linewidth]{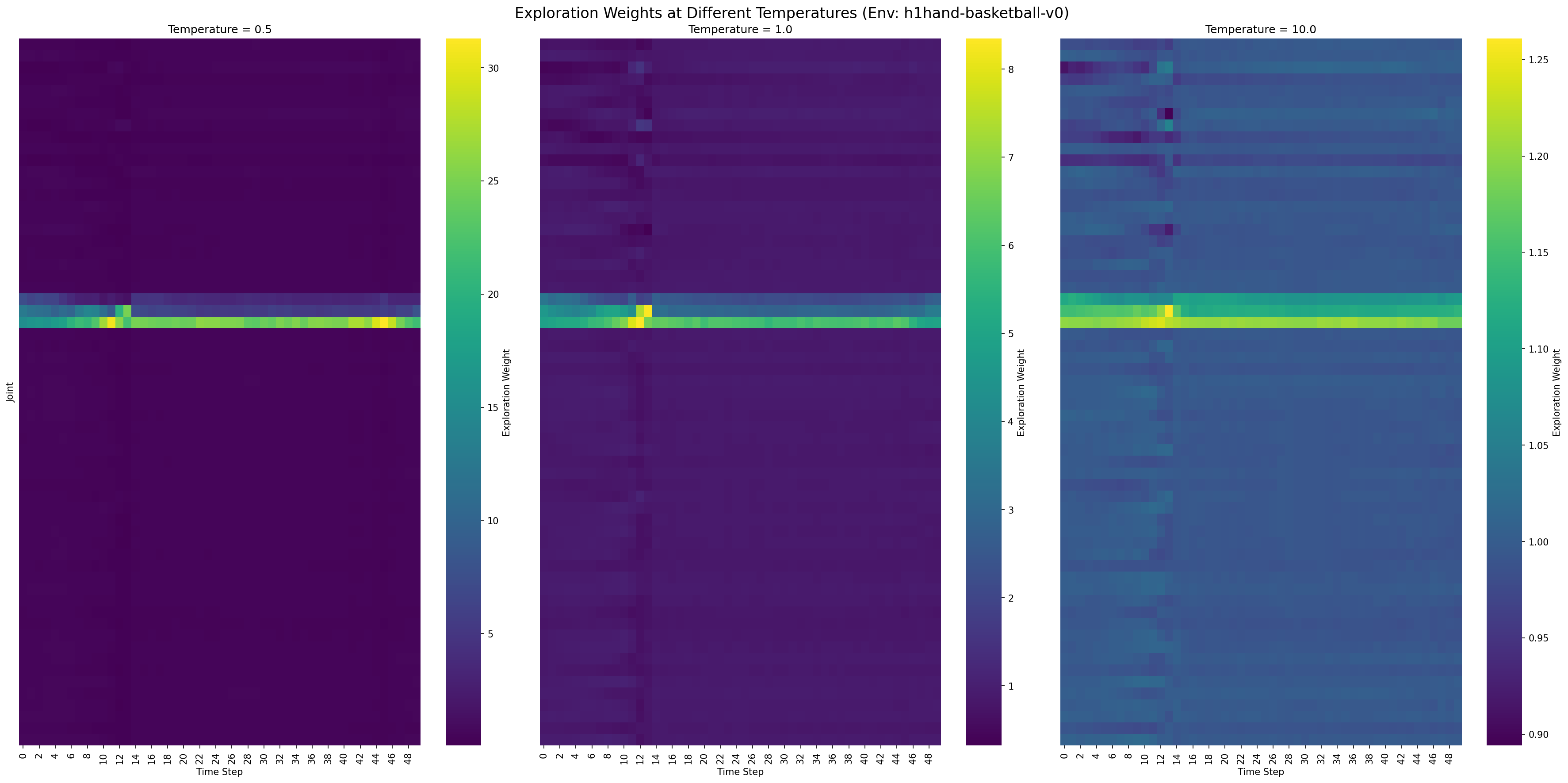}
	\caption{\textbf{Visualization of exploration weights at different temperatures.} Lower temperatures ($\tau=0.5$) result in highly concentrated weights on specific active joints, whereas higher temperatures ($\tau=10.0$) lead to a nearly uniform distribution across all joints.}
	\label{fig:tempcomps}
\end{figure}

\newpage
\section{C51 Support Sensitivity on \textit{Basketball}}
\label{app:c51_support}

\begin{figure}[H]
    \centering
    \includegraphics[width=0.85\linewidth]{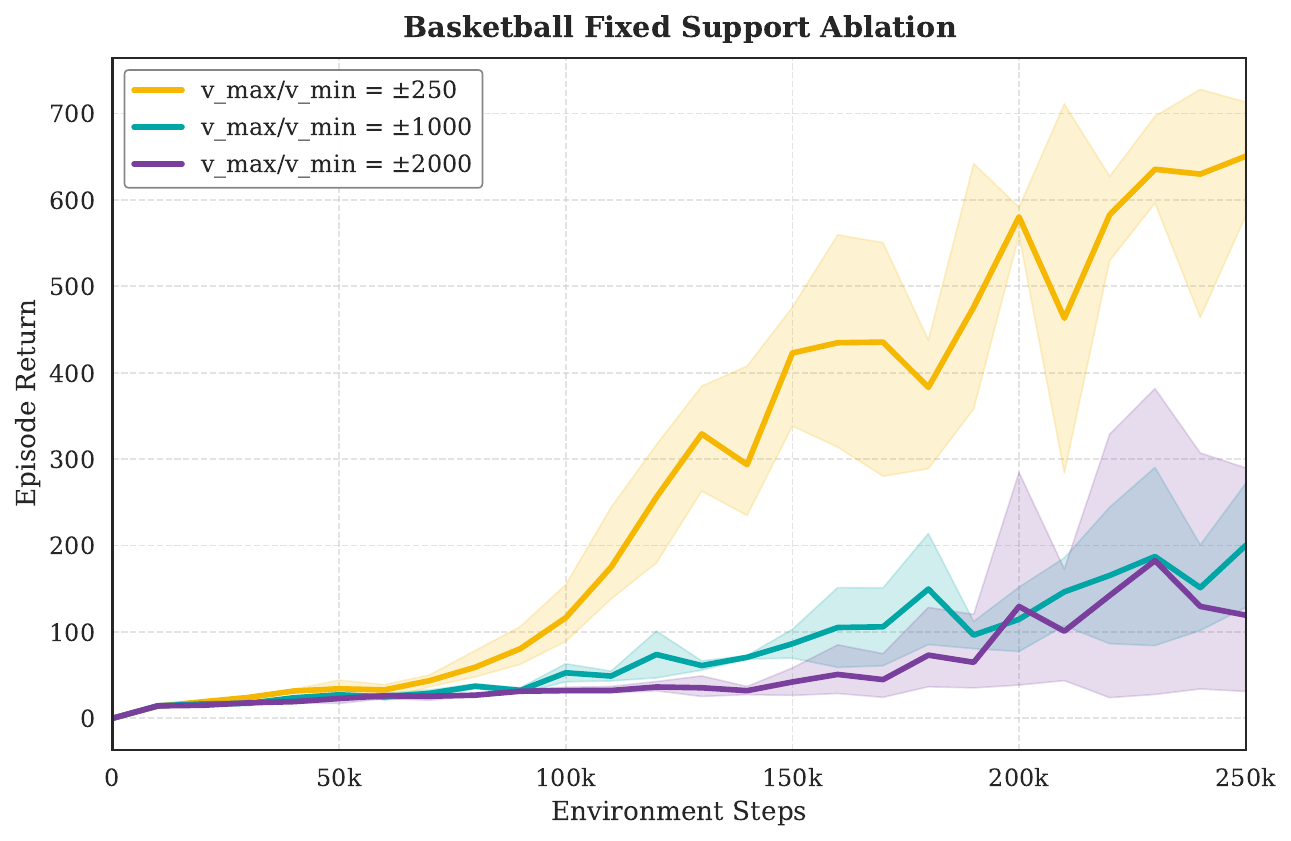}
    \caption{\textbf{Support sensitivity of C51$+$DEM on \textit{Basketball}.} We vary the C51 support range while keeping the remaining training pipeline unchanged. The resulting performance differences show that discrete critics can be sensitive to support specification on this task, whereas FastDSAC avoids this additional tuning axis.}
    \label{fig:c51_support_basketball}
\end{figure}

To directly examine the fixed-support issue for discrete critics, we varied the support range of C51$+$DEM on \textit{h1hand-basketball-v0} while keeping the actor architecture and the remaining training pipeline unchanged. As shown in Figure~\ref{fig:c51_support_basketball}, the performance of C51$+$DEM changes substantially with the support range: a narrower support achieves the best performance, while wider supports lead to noticeably worse results for \textit{Basketball} task. This sensitivity indicates that the support range constitutes an additional tuning axis for discrete critics in our setting. In contrast, FastDSAC avoids this support-specification burden by modeling the return distribution continuously.

\section{Ablation Study on Layer Normalization}
\label{sec:appendix_layernorm_ablation}

\begin{figure}[h]
    \centering
    \includegraphics[width=0.87\textwidth]{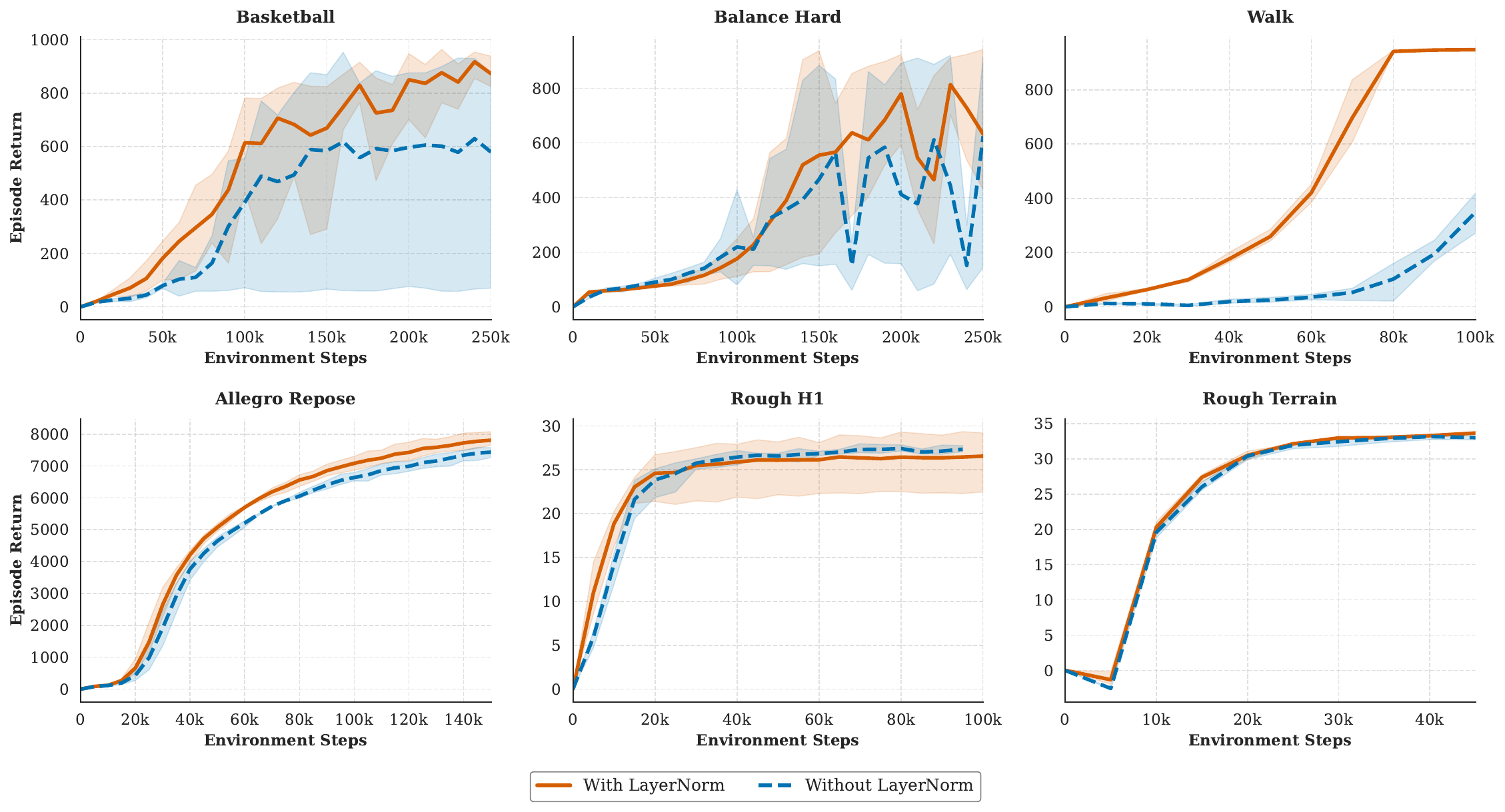}
    \caption{Ablation study of Layer Normalization. \textbf{Top row (HumanoidBench, $|\mathcal{A}|=61$):} Removing LayerNorm induces seed sensitivity, severe oscillations, and poor sample efficiency. \textbf{Bottom row (IsaacLab \& MuJoCo Playground, $|\mathcal{A}| < 30$):} FastDSAC remains highly stable and sample-efficient even without LayerNorm, indicating that the core algorithm is robust in lower-dimensional settings.}
    \label{fig:layernorm_ablation}
\end{figure}

To isolate the effect of Layer Normalization (LayerNorm) from our core algorithmic contributions, we conduct an ablation study on representative tasks from all three benchmarks, evaluating FastDSAC with and without LayerNorm. The results are shown in Figure~\ref{fig:layernorm_ablation}.

\textbf{Importance in High-Dimensional Settings (HumanoidBench).}
In the very high-dimensional domain ($|\mathcal{A}| = 61$), LayerNorm serves as a key architectural stabilizer. As shown in the top row of Figure~\ref{fig:layernorm_ablation}, removing LayerNorm substantially degrades training dynamics. Specifically, the agent exhibits high inter-seed variance in \textit{Basketball}, severe optimization oscillations and lower asymptotic returns in \textit{Balance Hard}, and drastically reduced sample efficiency in \textit{Walk}. These results suggest that, in expansive continuous action spaces, LayerNorm helps stabilize intermediate feature scales, leading to more reliable convergence and more sample-efficient learning in expansive continuous action spaces.

\textbf{Limited Impact in Lower-Dimensional Settings (IsaacLab \& MuJoCo Playground).} Conversely, in domains with moderate action dimensionality ($|\mathcal{A}| < 30$), the FastDSAC framework remains stable even without LayerNorm. As shown in the bottom row of Figure~\ref{fig:layernorm_ablation}, while adding LayerNorm yields marginal asymptotic improvements in certain manipulation tasks (e.g., \textit{Allegro Repose}), it has little effect on locomotion tasks overall, slightly degrading performance on \textit{Rough H1} while providing only a marginal improvement on \textit{Rough Terrain}. More importantly, even without LayerNorm, FastDSAC consistently outperforms the FastTD3 baselines in these domains (as established in Section 4.2). We therefore omit LayerNorm in these environments to highlight that FastDSAC's strong performance primarily stems from Dimension-wise Entropy Modulation (DEM) and the continuous distributional critic, rather than from architectural heuristics.

\section{Ablation Study on Target Entropy}
\label{sec:appendix_target_entropy_ablation}
\begin{wrapfigure}{r}{0.5\textwidth}\centering\includegraphics[width=0.8\linewidth]{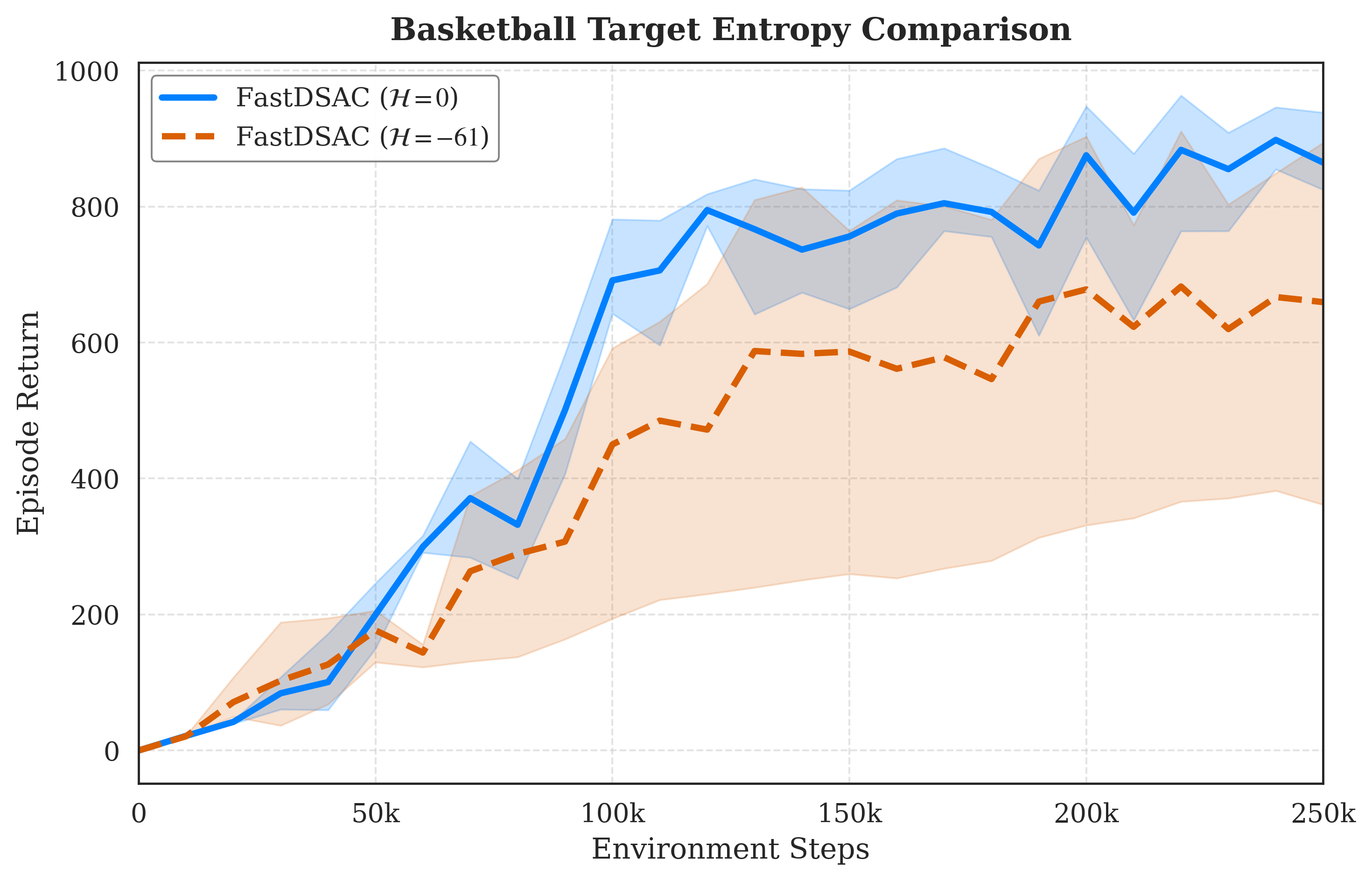}\caption{Ablation on target entropy in \textit{Basketball}. The standard heuristic ($\mathcal{H} = -61$) forces rapid variance decay, causing high inter-seed variance and premature convergence (orange curve). FastDSAC maintains $\mathcal{H} = 0$ to provide sufficient exploration capacity, which is structurally stabilized by DEM. Both curves report results over 3 seeds.}
\label{fig:entropy_ablation}
\end{wrapfigure}

In standard maximum entropy RL, the target entropy is typically set to $\mathcal{H} = -|\mathcal{A}|$ to regulate exploration. To evaluate our choice of $\mathcal{H} = 0$, we conduct an ablation study on the 61-DoF \textit{Basketball} task, comparing it against the standard heuristic ($\mathcal{H} = -61$).

\textbf{Premature Convergence under the Standard Setting.} As shown in Figure~\ref{fig:entropy_ablation}, enforcing the standard target ($\mathcal{H} = -61$) results in high inter-seed variance. While some seeds converge to near-optimal policies, others become prematurely trapped in local minima (converging to returns of approximately 300). This substantially degrades the average asymptotic performance, with final returns reaching only around 650. 

\textit{Analysis:} In a 61-DoF action space, $\mathcal{H} = -61$ induces rapid variance decay. As a result, the policy is forced to reduce its global exploration budget before identifying effective whole-body coordination patterns.

\textbf{Structural Management of High Entropy.} Setting $\mathcal{H} = 0$ prevents premature convergence by maintaining a sustained exploration budget throughout training. While such uniformly high variance would typically destabilize high-dimensional humanoid control, DEM manages this budget structurally. As analyzed in Section~\ref{sec:ablation:actor_ablation}, DEM prevents instability by offloading the excess variance onto redundant joints, preserving near-deterministic precision on critical actuators. Consequently, $\mathcal{H} = 0$ provides sufficient exploration capacity, and DEM ensures its stable allocation, yielding consistent and better convergence.

\section{Sensitivity to DEM Temperature}
\label{app:temperature_sensitivity}
\begin{figure}[H]
  \centering
  \includegraphics[width=\linewidth]{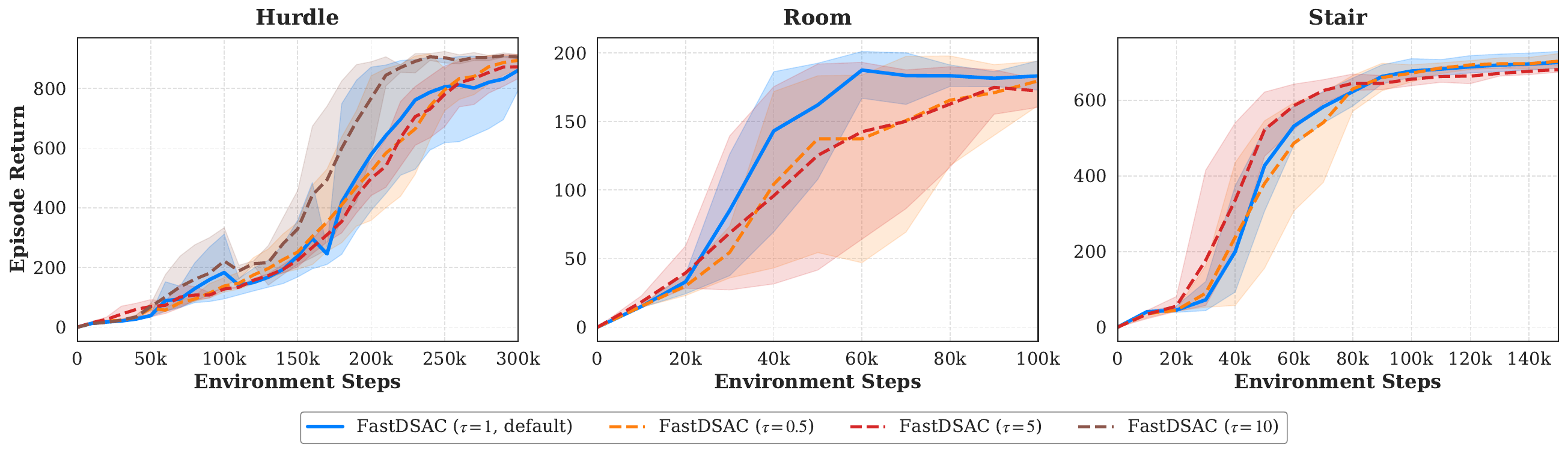}
  \caption{\textbf{Sensitivity to the DEM temperature $\tau$.} Learning curves on \textit{Hurdle}, \textit{Room}, and \textit{Stair} under different fixed temperatures. The default setting $\tau=1$
remains a robust choice overall. \textit{Room} favors $\tau=1$, \textit{Stair} shows similar performance across $\tau \in \{0.5,1,5\}$, and \textit{Hurdle} benefits from a larger temperature, with $
\tau=10$ achieving the best final return.}
  \label{fig:temperature_sensitivity}
\end{figure}

To examine how the fixed DEM temperature affects learning, we compare different $\tau$ values on three representative tasks: \textit{Hurdle}, \textit{Room}, and \textit{Stair} (Figure~\ref{fig:temperature_sensitivity}). For \textit{Room} and \textit{Stair}, we evaluate $\tau \in \{0.5, 1, 5\}$; for \textit{Hurdle}, we additionally include $\tau=10$.

The effect is clearly task-dependent. On \textit{Stair}, the three temperatures $\tau \in \{0.5,1,5\}$ all lead to similar final performance, indicating that this task is relatively insensitive to the precise sparsity level induced by DEM. On \textit{Room}, the default setting $\tau=1$ performs best, which aligns with our use of $\tau=1$ as the unified default configuration. On \textit{Hurdle}, however, a larger temperature is more effective: $\tau=10$ achieves the best final return, while the default $\tau=1$ is weaker than $\tau=0.5$, $\tau=5$, and $\tau=10$. This suggests that some highly dynamic coordination tasks can benefit from a less concentrated variance redistribution.

These observations motivate the FastDSAC (auto-$\tau$) variant. We indeed observe that end-to-end temperature optimization can further improve performance on some tasks. At the same time, since the fixed-$\tau$ configuration already performs strongly across the three benchmarks, we do not pursue this direction further here and instead leave a broader study of automatic temperature adaptation to future work.

\section{Full Results}
\label{app:full_results}
\begin{figure}[H]
    \centering
    \begin{subfigure}{\textwidth}
        \centering
        \includegraphics[width=0.93\textwidth]{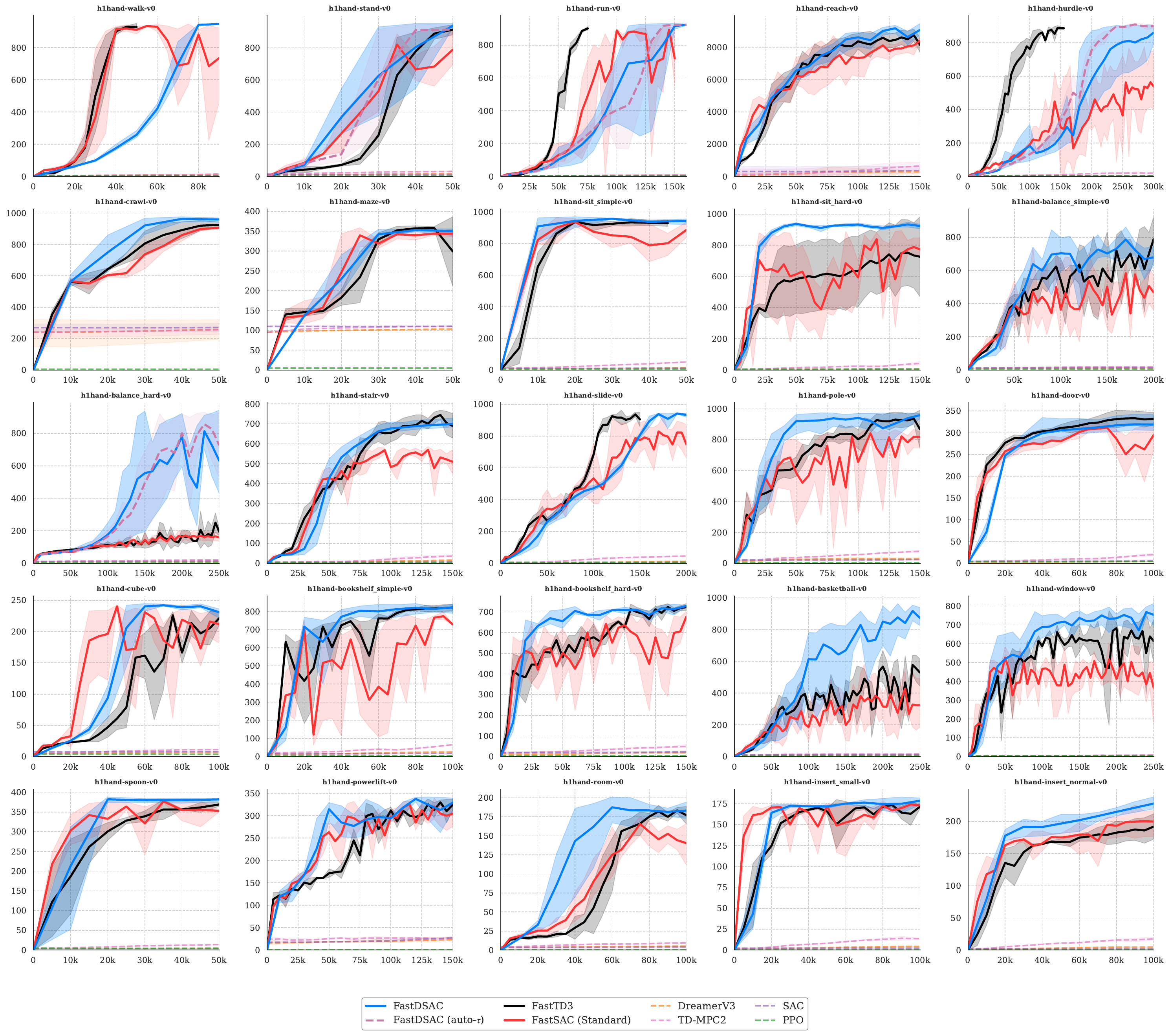}
        \caption{Full results on HumanoidBench (25 tasks).}
        \label{fig:full_results_hb}
    \end{subfigure}
    \begin{subfigure}{\textwidth}
        \centering
        \includegraphics[width=0.93\textwidth]{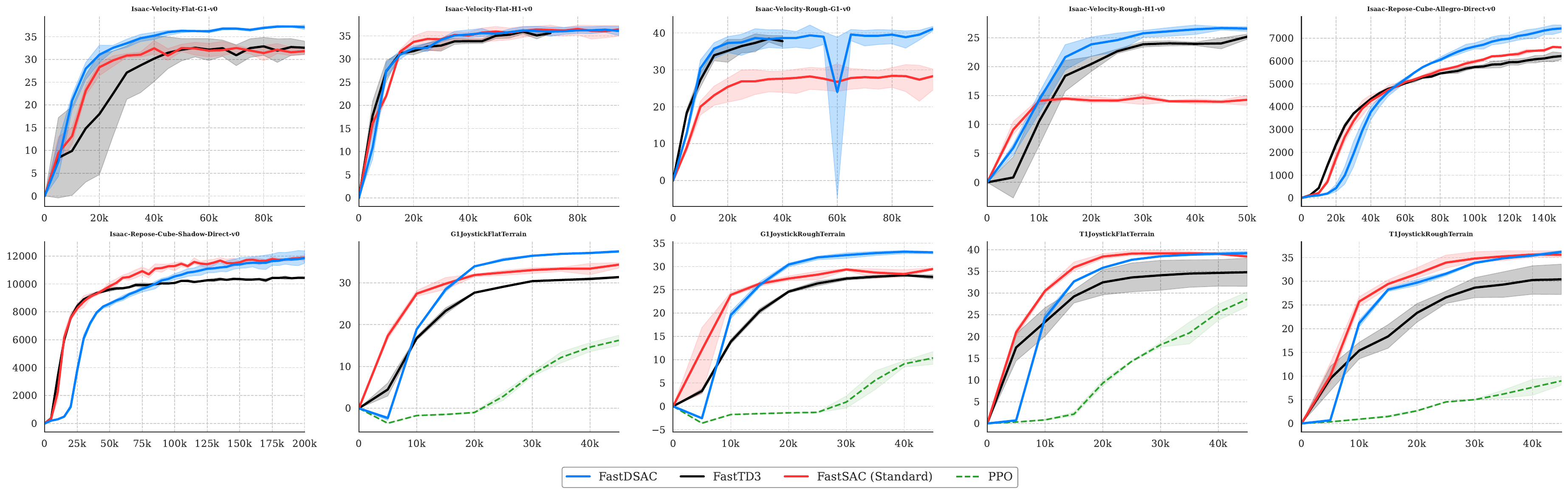}
        \caption{Full results on IsaacLab and MuJoCo Playground.}
        \label{fig:full_results_ilpg}
    \end{subfigure}
    \caption{\textbf{Full results across all benchmarks.} Curves show mean return over training steps with shaded regions indicating the min-max range across 3 seeds. \textbf{(a) HumanoidBench (25 tasks):} FastDSAC (blue) is compared against FastTD3 (black), FastSAC Standard (red), and leading baselines including DreamerV3 (orange), TD-MPC2 (pink), SAC (purple), and PPO (green). FastDSAC often outperforms or matches these baselines across locomotion (e.g., \textit{Run}, \textit{Hurdle}), fine manipulation (e.g., \textit{Insert}), and complex whole-body coordination (e.g., \textit{Basketball}, \textit{Balance}), combining strong exploration with stable training. \textbf{(b) IsaacLab and MuJoCo Playground (10 tasks):} FastDSAC (blue) is compared against FastTD3 (black), FastSAC Standard (red), and PPO (green). On IsaacLab, FastDSAC performs strongly on locomotion (Velocity Flat/Rough H1/G1) and Repose Cube Allegro, while remaining competitive on Repose Cube Shadow. On MuJoCo Playground, FastDSAC consistently outperforms baselines on G1/T1 Joystick with lower variance and higher asymptotic returns.}
    \label{fig:full_results_all}
\end{figure}

\section{Discussion of Limitations, Future Directions and Impact}
\label{sec:limitations}
\paragraph{Limitations.} 
The primary limitation of FastDSAC is that Dimension-wise Entropy Modulation (DEM) assumes independent exploration noise across action dimensions (i.e., utilizing a diagonal covariance matrix). While this formulation effectively scales to 61-DoF systems, it precludes the modeling of correlated exploration strategies among interdependent joints. Furthermore, maintaining a continuous distributional critic to ensure accurate value estimation comes at the cost of a marginal increase in computational overhead during training compared to purely deterministic baselines.
\paragraph{Future Directions.}
An important direction for future work is to study how structured covariance models might be used to capture joint dependencies, and how end-to-end optimization of DEM's temperature parameter can enable fully autonomous variance scheduling.
\paragraph{Impact Statements.}
This paper presents work whose goal is to advance the field of Deep Reinforcement Learning.
The deployment of such agile autonomous systems raises valid safety concerns regarding unintended physical interaction and potential long-term economic impacts on the workforce. Consequently, real-world application of these policies necessitates robust verification protocols and rigorous safety constraints to prevent accidents and ensure physical safety.
\clearpage
\section*{NeurIPS Paper Checklist}

\begin{enumerate}

\item {\bf Claims}
    \item[] Question: Do the main claims made in the abstract and introduction accurately reflect the paper's contributions and scope?
    \item[] Answer: \answerYes{} 
    \item[] Justification: The abstract and introduction state the same core contributions that are later supported by the experiments and conclusion: DEM, the continuous distributional critic, strong results on 35 tasks, and zero-shot sim-to-real transfer (Abstract, Introduction, Sections \ref{sec:main_results}--\ref{sec:sim2real}, Conclusion).

\item {\bf Limitations}
    \item[] Question: Does the paper discuss the limitations of the work performed by the authors?
    \item[] Answer: \answerYes{} 
    \item[] Justification: The paper includes a dedicated limitations section in Section~\ref{sec:limitations}. It explicitly notes the diagonal-covariance assumption in DEM and the marginal training overhead of the continuous critic, and it also points to structured covariance models and auto-$\tau$ as future directions. 

\item {\bf Theory assumptions and proofs}
    \item[] Question: For each theoretical result, does the paper provide the full set of assumptions and a complete (and correct) proof?
    \item[] Answer: \answerNA{} 
    \item[] Justification: The paper does not present proofs; the convergence statement is attributed to prior DSAC/DSAC-T work, while the rest of the paper focuses on algorithm design and experiments.

    \item {\bf Experimental result reproducibility}
    \item[] Question: Does the paper fully disclose all the information needed to reproduce the main experimental results of the paper to the extent that it affects the main claims and/or conclusions of the paper (regardless of whether the code and data are provided or not)?
    \item[] Answer: \answerYes{} 
    \item[] Justification: The paper provides: (1) an anonymous implementation link in the Introduction; (2) complete hyperparameter configurations in Appendix~\ref{app:hyperparameters}; (3) benchmark/task lists, training protocols, seed counts, and evaluation settings in Section~\ref{sec:experiments} and Appendix~\ref{app:hyperparameters}; (4) sim-to-real deployment details in Appendix~\ref{app:sim2real}. All baselines use public codebases with configuration details specified. 

\item {\bf Open access to data and code}
    \item[] Question: Does the paper provide open access to the data and code, with sufficient instructions to faithfully reproduce the main experimental results, as described in supplemental material?
    \item[] Answer: \answerYes{} 
    \item[] Justification: The paper provides an anonymous implementation link and describes how the benchmarks and evaluation pipelines are configured. The experiments rely on public benchmark environments rather than proprietary datasets, and the appendix gives the settings needed to run them (Introduction, Appendix \ref{app:hyperparameters}).

\item {\bf Experimental setting/details}
    \item[] Question: Does the paper specify all the training and test details (e.g., data splits, hyperparameters, how they were chosen, type of optimizer) necessary to understand the results?
    \item[] Answer: \answerYes{} 
    \item[] Justification: Section \ref{sec:experiments} and Appendix \ref{app:hyperparameters} specify the benchmarks, baselines, seed counts, parallel environment counts, replay/update schedules, discount factors, target entropy, and benchmark-specific architectural choices such as LayerNorm. 

\item {\bf Experiment statistical significance}
    \item[] Question: Does the paper report error bars suitably and correctly defined or other appropriate information about the statistical significance of the experiments?
    \item[] Answer: \answerYes{} 
    \item[] Justification: The main figures report mean returns over 3 seeds with shaded min-max ranges, and the hardest tasks use 5 seeds with shaded regions spanning the min-max range across seeds, so the paper does provide a clear measure of run-to-run variability for the key results (Sections \ref{sec:experiments}, \ref{sec:main_results}, Appendix \ref{app:full_results}).

\item {\bf Experiments compute resources}
    \item[] Question: For each experiment, does the paper provide sufficient information on the computer resources (type of compute workers, memory, time of execution) needed to reproduce the experiments?
    \item[] Answer: \answerYes{} 
    \item[] Justification: The Computational Resources section and Appendix \ref{app:wall_clock_time} specify the CPU/GPU/RAM configurations, task allocation across hardware, and per-task as well as total wall-clock time needed for the representative experiments. 
    
\item {\bf Code of ethics}
    \item[] Question: Does the research conducted in the paper conform, in every respect, with the NeurIPS Code of Ethics \url{https://neurips.cc/public/EthicsGuidelines}?
    \item[] Answer: \answerYes{} 
    \item[] Justification: The work uses public simulation benchmarks and a robot deployment, and the paper does not describe any practice that appears to conflict with the NeurIPS Code of Ethics. No human-subject data, privacy-sensitive data, or prohibited collection practices are involved. 

\item {\bf Broader impacts}
    \item[] Question: Does the paper discuss both potential positive societal impacts and negative societal impacts of the work performed?
    \item[] Answer: \answerYes{} 
    \item[] Justification: See Appendix~\ref{sec:limitations}. 
    
\item {\bf Safeguards}
    \item[] Question: Does the paper describe safeguards that have been put in place for responsible release of data or models that have a high risk for misuse (e.g., pre-trained language models, image generators, or scraped datasets)?
    \item[] Answer: \answerNA{} 
    \item[] Justification: The paper does not release a high-risk dual-use model or scraped dataset, so there are no special safeguard procedures to describe beyond the standard anonymous implementation release. 

\item {\bf Licenses for existing assets}
    \item[] Question: Are the creators or original owners of assets (e.g., code, data, models), used in the paper, properly credited and are the license and terms of use explicitly mentioned and properly respected?
    \item[] Answer: \answerYes{} 
    \item[] Justification: The paper uses open-source benchmarks (HumanoidBench, MuJoCo Playground, IsaacLab) and the FastTD3 codebase. All assets are properly cited in the text with their original papers. HumanoidBench is under MIT License, MuJoCo Playground under Apache 2.0, IsaacLab under BSD 3-Clause, and FastTD3 is publicly available. The Unitree G1 robot is a commercially available platform. 

\item {\bf New assets}
    \item[] Question: Are new assets introduced in the paper well documented and is the documentation provided alongside the assets?
    \item[] Answer: \answerYes{} 
    \item[] Justification: The new asset here is the anonymous implementation of FastDSAC, and it is documented through the method section, Algorithm \ref{alg:fastdsac}, the hyperparameter tables, and the sim-to-real appendix. 

\item {\bf Crowdsourcing and research with human subjects}
    \item[] Question: For crowdsourcing experiments and research with human subjects, does the paper include the full text of instructions given to participants and screenshots, if applicable, as well as details about compensation (if any)? 
    \item[] Answer: \answerNA{} 
    \item[] Justification: The paper does not involve crowdsourcing or any human-subject study; the real-world evaluation is on a humanoid robot, not on participants. 

\item {\bf Institutional review board (IRB) approvals or equivalent for research with human subjects}
    \item[] Question: Does the paper describe potential risks incurred by study participants, whether such risks were disclosed to the subjects, and whether Institutional Review Board (IRB) approvals (or an equivalent approval/review based on the requirements of your country or institution) were obtained?
    \item[] Answer: \answerNA{} 
    \item[] Justification: Because the paper does not collect data from human participants or run a human-subject study, IRB approval is not applicable. 

\item {\bf Declaration of LLM usage}
    \item[] Question: Does the paper describe the usage of LLMs if it is an important, original, or non-standard component of the core methods in this research? Note that if the LLM is used only for writing, editing, or formatting purposes and does \emph{not} impact the core methodology, scientific rigor, or originality of the research, declaration is not required.
    \item[] Answer: \answerNA{} 
    \item[] Justification: The core method, experiments, and results do not rely on LLMs, and LLMs were only used for polishing and formatting purposes; the paper's scientific contributions are entirely in reinforcement learning and robotics. 

\end{enumerate}

\end{document}